\RequirePackage{fix-cm}
\documentclass[twocolumn]{svjour3}          %
\smartqed  %
\usepackage{graphicx}
\usepackage{amsmath}
\usepackage{multirow}
\usepackage{pifont}
\usepackage{graphicx}
\usepackage{amssymb}
\usepackage{array}
\usepackage{multirow, nicefrac}
\usepackage{pifont}
\usepackage{cite}
\usepackage{bbm}
\usepackage{arydshln}
\usepackage[dvipsnames]{xcolor}
\usepackage{tablefootnote}
\usepackage{threeparttable}
\definecolor{green}{rgb}{1,0,0}
\usepackage{stfloats}
\usepackage{color}
\usepackage[normalem]{ulem}
\usepackage[switch]{lineno} 
\usepackage[symbol]{footmisc}
\usepackage{tabularx}
\usepackage{lineno}

\usepackage{times}
\usepackage{epsfig}
\usepackage{float}
\usepackage{color}
\usepackage{bbold}
\usepackage[ruled,linesnumbered]{algorithm2e}
\usepackage{booktabs}
\usepackage{ragged2e}
\usepackage[pagebackref=true,breaklinks=true,letterpaper=true,colorlinks,bookmarks=false]{hyperref}
\newcommand{\tabincell}[2]{\begin{tabular}{@{}#1@{}}#2\end{tabular}}

\def\0{{\bf 0}}
\def\1{{\bf 1}}

\usepackage{booktabs}
\usepackage[utf8]{inputenc}

\DeclareRobustCommand\onedot{\futurelet\@let@token\@onedot}
\def\@onedot{\ifx\@let@token.\else.\null\fi\xspace}

\makeatother

\usepackage{listings,xcolor} 
\usepackage{tcolorbox}
\newcommand{\lstfont}[1]{\color{#1}\ttfamily}

\usepackage{xspace}

\usepackage{marvosym}

\makeatletter
\def\cl@chapter{\@elt {theorem}}
\makeatother
\usepackage[capitalise]{cleveref}

\begin{document}
\sloppy
\title{Identity-free Artificial Emotional Intelligence via Micro-Gesture Understanding
}

\author{
Rong Gao\footnotemark[1],
        Xin Liu\footnotemark[1] \footnotemark[2],
        Bohao Xing,
        Zitong Yu,
        Bjorn W. Schuller,
        and Heikki Kälviäinen %
}
\authorrunning{Rong Gao, Xin Liu, Bohao Xing, Zitong Yu, Bjorn W. Schuller, Heikki Kälviäinen} %

\institute{
Rong Gao, Xin Liu, Bohao Xing, and Heikki Kälviäinen \at
              Computer Vision and Pattern Recognition Laboratory\\ School of Engineering Sciences\\ Lappeenranta-Lahti University of Technology LUT \\
              Finland\\
              \email{\{rong.gao, xin.liu, bohao.xing,  heikki.kalviainen\}@lut.fi}
           \and
           Zitong Yu \at
           School of Computing and Information Technology\\
           Great Bay University\\ 
           China\\
            \and
           Bjorn W. Schuller \at  
           Group on Language, Audio, $\&$ Music\\ 
           Imperial College London\\ 
           United Kingdom\\ 
           and\\
           School of Medicine and Health\\
           Technical University of Munich\\
           Germany\\
              }

\date{\today}

\maketitle

\footnotetext[1]{These authors contributed equally to this work.}
\footnotetext[2]{Corresponding author.}

\maketitle

\begin{abstract}
In this work, we focus on a special group of human body language --- the {\bf{micro-gesture (MG)}}, which differs from the range of ordinary illustrative gestures in that they are not intentional behaviors performed to convey information to others, but rather unintentional behaviors driven by inner feelings. This characteristic introduces two novel challenges regarding micro-gestures that are worth rethinking. The first is whether strategies designed for other action recognition are entirely applicable to micro-gestures. The second is whether micro-gestures, as supplementary data, can provide additional insights for emotional understanding.  In recognizing micro-gestures, we explored various augmentation strategies that take into account the subtle spatial and brief temporal characteristics of micro-gestures, often accompanied by repetitiveness, to determine more suitable augmentation methods. Considering the significance of temporal domain information for micro-gestures, we introduce a simple and efficient plug-and-play spatiotemporal balancing fusion method. We not only studied our method on the considered micro-gesture dataset but also conducted experiments on mainstream action datasets. The results show that our approach performs well in micro-gesture recognition and on other datasets, achieving state-of-the-art performance compared to previous micro-gesture recognition methods. For emotional understanding based on micro-gestures, we construct complex emotional reasoning scenarios. Our evaluation, conducted with large language models, shows that micro-gestures play a significant and positive role in enhancing comprehensive emotional understanding. We confirm that our new insights contribute to advancing research in micro-gesture and emotional artificial intelligence. Our code is available at: \href{https://github.com/Erich-G/MG_based_Emotion_Understanding}{https://github.com/Erich-G/MG\_based\_Emotion\_Understanding}.

\keywords{Micro-gesture Recognition \and Emotion Understanding \and Contrastive Learning \and Large Language Model 
}
\end{abstract}

\section{Introduction}

\begin{figure}[htp]
\begin{center}
\includegraphics[width=8.2cm]{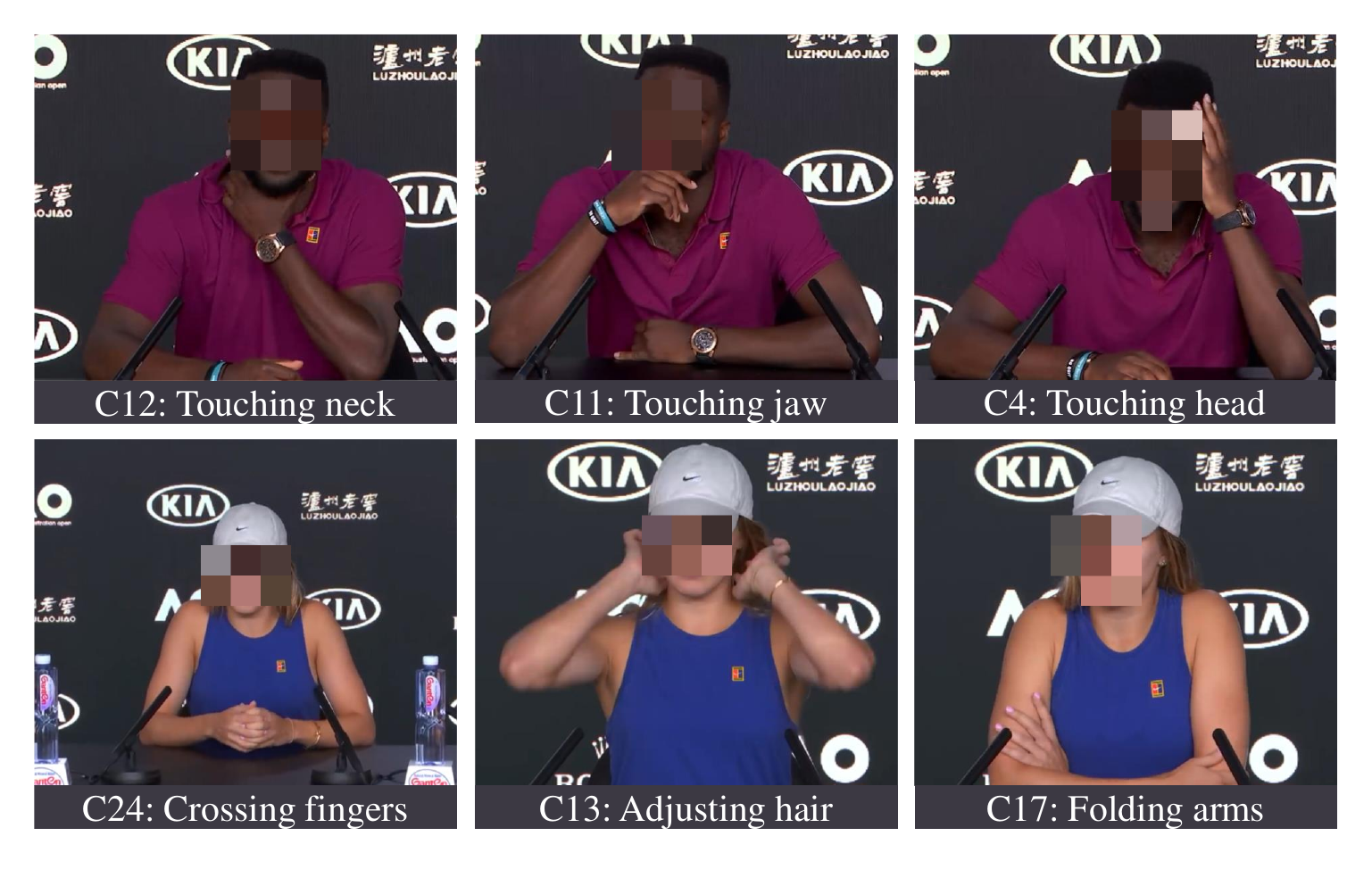}
\end{center}
\vspace{-0.5cm}
\caption{Frames from post-match press conference videos exhibit various identity-free micro-gestures (MGs). The objective of this study is to empower machines to recognize these MGs, comprehend the players' emotional states holistically, and subsequently determine whether the player has won or lost the match (positive or negative emotional states).}
\vspace{-0.6cm}
\label{fig:intro-lose&win}
\end{figure}
\setlength{\belowcaptionskip}{10pt}

Artificial Emotional Intelligence, also referred to as Emotion AI, is the ability of computers to understand human emotional states. It is a critical component in human-machine communication, as emotions are constantly present in human behavior, thinking, and decision-making. Every day, we encounter various emotional cues, including facial expressions, eye movements, tone of voice, gestures, posture, touch, and proximity. These cues, such as smiling or waving, can reveal our true selves and shape our relationships with others, making Emotion AI a rapidly growing field of interest. A primary area of focus within this field is the analysis of the human face \cite{kanade2000comprehensive,pantic2005web, yin20063d, valstar2010induced,gross2010multi,zhang2013high, aung2015automatic,lucey2011painful,bartlett2006fully,mckeown2011semaine,dhall2017individual,kollias2019deep, yan2014casme,li2013spontaneous,davison2016samm}. Particularly, facial expressions play a significant role in Emotion AI, as they are widely acknowledged as a crucial means of conveying emotions among humans. People have long used facial expressions to read and understand emotions. Eye gaze is another major form of facial behavior, as eye movements, such as looking, staring, and blinking, are important nonverbal signals. Increased blinking frequency and pupil dilation often occur when people encounter things or people they like. Glancing at another person can communicate various emotions, including hostility, interest, and attraction. The tone of voice is also a vocal signal that has received significant attention \cite{schuller2011avec, schuller2012avec, mckeown2011semaine, ringeval2013introducing, joo2015panoptic,joo2019towards} in Emotion AI. For example, speaking in a confident tone conveys approval and enthusiasm to the audience, while speaking in a hesitant tone conveys disapproval and disinterest. Recently, multi-modal emotion benchmarks \cite{soleymani2011multimodal,koelstra2011deap,mckeown2011semaine,ringeval2013introducing} have been developed, covering not only facial and speech signals, but also electrocardiogram (ECG), electroencephalogram (EEG), and galvanic skin response data. It is important to note that these datasets contain sensitive biometric information and may face increasing restrictions due to privacy concerns. Furthermore, these benchmarks concentrate primarily on identifying and classifying emotional behaviors singularly, whereas the objective of Emotion AI is to comprehend emotions that lie beyond these behaviors. 
With these limitations in mind, in previous research, we presented an identity-free video dataset for Micro-Gesture Understanding and Emotion analysis (iMiGUE) \cite{Liu_2021_CVPR} that differs from previous datasets with several unique features. 

iMiGUE is the first publicly available dataset for studying emotions through micro-gestures (MGs) in a privacy-preserving manner. Unlike previous studies that focused on illustrative (or intentional) behaviors \cite{vinciarelli2009social}, e.g., waving hands as a greeting. However, on many occasions, people would suppress or hide their feelings (especially negative ones) rather than express them. Psychological studies \cite{aviezer2012body,axtell1991gestures,burgoon1989nonverbal} showed that there is a special group of gestures, the micro-gestures. The major difference between MGs and illustrative gestures is that MGs are unintentional behaviors elicited by inner feelings, e.g., rubbing one's hands due to stress. The function of MGs is to relieve or protect oneself from negative feelings rather than presenting them to others. In \cite{ekman2009telling}, the authors showed that there are over 215 behaviors associated with psychological discomfort and most of them are not only in the face. MGs are subtle and some of them are short, mostly out of our awareness or notice during live interactions \cite{ginger2007gestalt}. It would be of great value if we could develop machine learning methods to capture and recognize these neglected clues for better emotional understanding, and iMiGUE provides fertile ground for this. Especially, most of the existing studies on Emotion AI are highly correlated with sensitive biometric data. Compared to the biometric data about the most intimate nuances of people’s real-world faces and speech, the identity-free MGs from user gestural or other bodily expressions provide a safe way to analyze human emotions.

iMiGUE spans the gap between behavioral recognition and emotional understanding. Most datasets focus only on recognizing behaviors, but not the emotions behind them. iMiGUE bridges this gap by including both MGs and emotions in its annotations. Let us consider a post-game interview scenario where a player is being interviewed 
as seen in Fig.~\ref{fig:intro-lose&win}. One may observe MGs such as playing with hair (relaxation) or touching the neck or head (tension or chagrin), but to truly understand the player's emotions (positive or negative, caused by winning or losing the game), a machine needs to recognize these behaviors at both the micro-gesture and emotion levels. iMiGUE is designed to do just that, 
not only recognize MGs but also provide a hierarchical understanding of their relationship to emotions.


\begin{figure*}[t]
\begin{center}
\includegraphics[width=18cm]{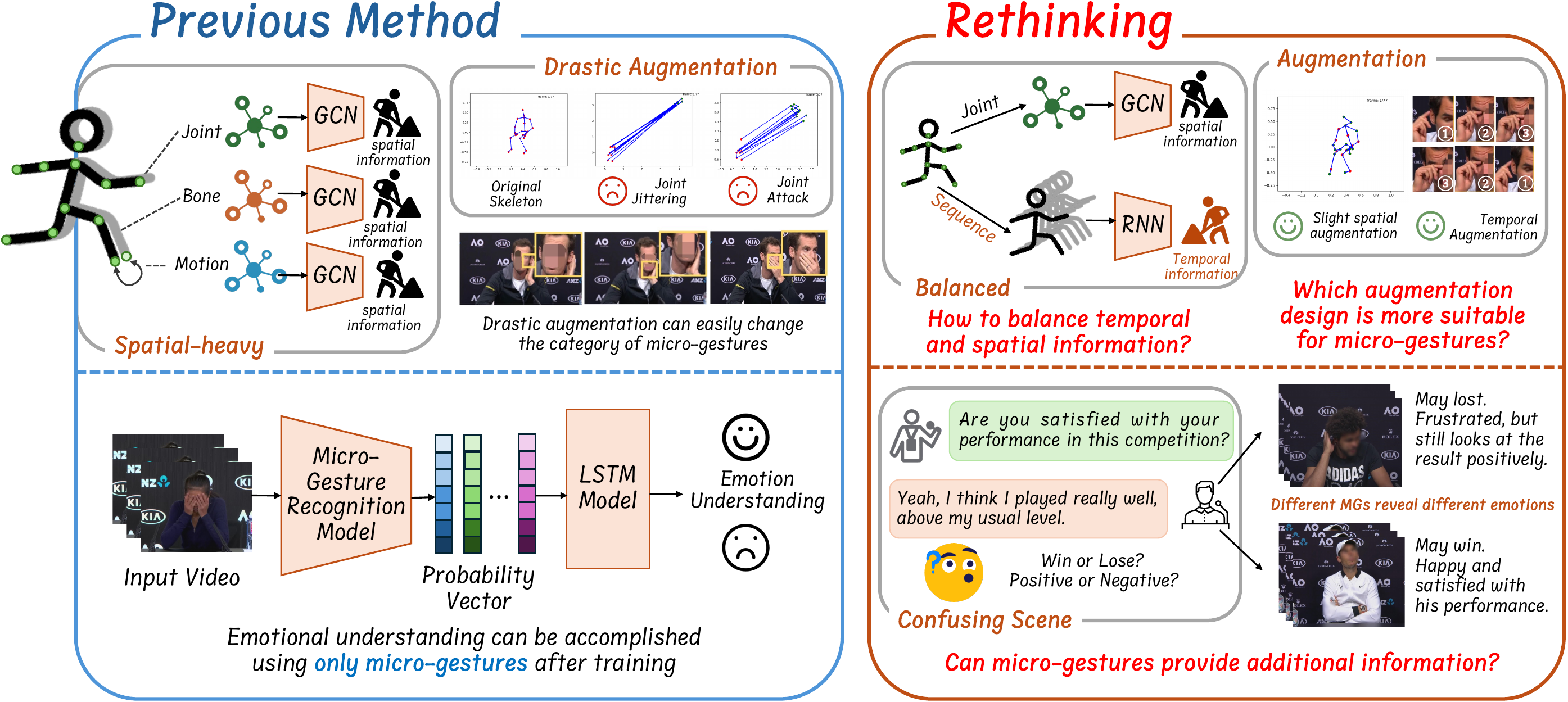}
\end{center}
\caption{Previous methods mostly focused on spatial feature mining, and the augmentation techniques used were not designed with the characteristics of micro-gestures in mind. Therefore, it is worth carefully considering how to balance temporal and spatial information and exploring augmentation methods suitable for micro-gestures. Our previous research \cite{Liu_2021_CVPR} has demonstrated to some extent the effectiveness of using only micro-gestures for emotional understanding. Now, let us consider more realistic scenarios: can micro-gestures, as an auxiliary role, provide additional effective information?}
\label{fig:re_think}
\end{figure*}

In the preliminary version of this work \cite{Liu_2021_CVPR}, we evaluated various existing methods on the iMiGUE dataset and found that traditional supervised learning techniques did not yield satisfactory results. To address this issue, we proposed an unsupervised sequential variational autoencoder (U-S-VAE), an unsupervised encoder-decoder-based approach. 
However, this model lacks a semantic connection between the encoder and the decoder, hindering the extraction of meaningful features. In light of the recent success of Contrastive Learning in representation learning, we aim to utilize this approach to make a breakthrough. Additionally, in the initial version, we used micro-gestures alone as cues for understanding emotions. The results showed that it is somewhat possible to infer underlying emotions solely from micro-gestures. However, most real-world scenarios are likely more complex, and relying solely on micro-gestures for inference may not be sufficient for these complex situations. Based on these considerations, we propose two new reevaluations of iMiGUE as illustrated in Fig.~\ref{fig:re_think}. 1) Due to the lack of research on micro-gestures, it is worth exploring whether augmentation strategies designed for action recognition in other works are still applicable to micro-gestures. Additionally, previous work has mostly focused on mining spatial information, which may be due to the larger spatial differences in sports and everyday expressive gestures, while micro-gestures, influenced by their temporal characteristics, require more balanced spatio-temporal feature mining. 2) In previous research, we explored the task of emotion inference based purely on micro-gestures. However, in a wild environment, micro-gestures rarely occur in isolation. They are often accompanied by other modalities of information, such as micro-expressions, speech, etc. Imagine a scenario where a journalist interviews an athlete, asking if they are satisfied with their performance in the competition. The athlete responds, ``Yes, I think I performed very well, beyond my usual level." Can we determine their success or failure based solely on these words? If additional micro-gesture information is provided, such as one athlete scratching their head or covering their face while saying this, and another standing upright and lifting their head, can these details provide effective incremental information for the final inference? Validating this question plays a crucial role in determining the direction of further research on micro-gestures.

There are two important factors that must not be over-looked to cope with the first rethink -  distinguishing between illustrative gestures and MGs: MGs differ from illustrative gestures (actions) in that they are localized in certain areas of the body and characterized by smaller movements, limited to a few joints. Unlike illustrative gestures, MGs are not consciously controlled and exhibit more individuality, with no conventional norms. They can be fleeting or repetitive and display unique spatial and temporal properties, including ``tiny'' and ``individual'' spatial characteristics and ``short'' and ``repetitive'' temporal characteristics. As an important module in contrastive learning, the augmentation methods designed for illustrative behaviors in previous action recognition research need to be improved to make them more suitable for micro-gestures. Additionally, some new augmentation methods are worth exploring. As to the importance of temporal information for MGs, the temporal semantics of MGs are crucial to their interpretation. However, the mainstream approach which incorporated joint, bone, and motion information into the ``graph'' structure still has a limitation of being heavily focused on the spatial aspect. As a result, this method only enhances the temporal information within the ``graph'' modality and fails to fully exploit the rich temporal information contained in the ``sequence'' modality. To achieve a more comprehensive understanding of MGs, a better balance between temporal and spatial features is required.

To address the limitations mentioned above, we explore improved augmentation methods for MGs and introduce a new Spatial-Temporal-Balanced Dual-Stream Fusion network. To deal with the second rethink, we constructed a complex scenario and evaluated whether micro-gestures can serve as an auxiliary role to provide additional information for reasoning. Considering the extensive knowledge boundaries of large language models (LLMs), we employ large language models to act as subjects in this complex scenario, as psychology experts with certain knowledge in psychology, tennis events, and an understanding of the relationship between psychology and micro-gestures to some extent. This paper is an extension of our earlier publication  \cite{Liu_2021_CVPR}
. Our enhancements are in the methodology and experiments. Here, we present several key differences
as follows:
1) A new dual-stream contrastive learning model is proposed to replace the U-S-VAE that effectively integrates temporal and spatial information. 2) New augmentation methods specifically explored for better MGs contrastive learning are employed. 3) A complex emotional reasoning scenario is constructed to verify whether micro-gestures can provide additional effective information as an auxiliary role. 4) We expand the experiments and present results on additional datasets, more comprehensive comparisons to other models, and ablation studies. The key contributions of this paper can be summarized as follows:

\begin{itemize}

\item We explore augmentation methods tailored to the MGs. These methods allow for the augmented data to not only be semantic and meaningful, but also to extensively explore the boundaries of the micro-gesture category. In particular, we explore ``coordinate perturbation'' and ``viewpoint perturbation'' to address the ``smallness/micro'' of MGs, ``stretching'' and ``posterize time'' for the ``uniqueness'' of each MG, and ``reverse'' and "repeat'' for the ``repetitive'' nature of MGs.

\item We explore a dual-stream contrastive learning network to provide a comprehensive understanding of MGs. In the temporal stream, we organize the data in the form of sequences. In the spatial stream, we manage the data in the form of graphs.

\item  
We improve and expand the experiments.
The results show that even the state-of-the-art (SOTA) fully supervised learning model does not achieve satisfactory accuracy on the iMiGUE dataset, which highlights the difficulty of identifying such subtle MGs. However, the proposed method significantly outperforms baseline methods on the iMiGUE dataset. To further validate the effectiveness of our model, we conduct experiments on multiple benchmarks, including the NTU RGB+D 60 and NTU RGB+D 120 datasets. Experimental results demonstrate the superiority of our approach.

\item We employed LLMs to evaluate the role of micro-gestures in assisting emotional understanding. The complex sense we designed can be extended to other tasks related to micro-gestures, such as deception detection and interviews. Experimental results demonstrate that micro-gestures play a positive role as an aid in emotion understanding. This lays a foundation for our subsequent research.

\end{itemize}
\section{Related Work}
Body language can convey a person's emotional state, so analyzing physical activities such as actions, gestures, and postures are the popular research topics \cite{soomro2012ucf101,kuehne2011hmdb, gorban2015thumos,rohrbach2016recognizing,shahroudy2016ntu,sigurdsson2016hollywood, goyal2017something,monfort2019moments,yu2020humbi} in the community. These datasets then focus on identifying human activities and rarely make further correlations with emotional states. In this section, we provide a review of relevant emotional gesture-based benchmarks. Then, we review related work on gesture/action recognition.

\subsection{Emotional Gesture/Action-based Datasets.}
As one of the critical social communication cues, gestures encompass the movement of the hands, head, and other parts of the body and can be a good expression of inner feelings, emotions, and thoughts \cite{noroozi2018survey}. The attributes of the currently widely used emotional gesture database are displayed in Table~\ref{tb:datasetscom}. Most of the early studies in this area were based on behavioral or postural gestures. The stimulus set \cite{schindler2008recognizing} from Tilburg University collects photographs of 50 actors performing different emotions. The FABO database \cite{gunes2006bimodal} is a pioneering work, and the use of video clips of intentionally posed prototypical gestures for emotion recognition is proposed in this work. The videos in the dataset were tagged with six basic emotions and four states: neutral, anxiety, boredom, and uncertainty. Inspired by this research on the behavior of filmed poses under controlled recording conditions, emotional gesture analysis was gradually extended to many directions. In HUMAINE \cite{douglas2007humaine,castellano2007recognising}, the researchers elicited emotions by interacting with the computer avatar of the operator. The Geneva multi-modal emotion portrayals (GEMEP) database \cite{glowinski2008technique} contains more than 7\,000 audio-video portrayals of 18 emotions portrayed by 10 actors. Also, in a controlled environment, the subset \cite{gavrilescu2015recognizing} of LIRIS-ACCEDE database \cite{baveye2015liris}, collected upper body emotional gestures from 64 subjects. Using the Kinect sensor, Saha \textit{et al.} \cite{saha2014study} collected 3D skeletal gesture data from 10 subjects labeled with five evoked emotions: anger, fear, happiness, sadness, and relaxation. Psaltis \textit{et al.} \cite{psaltis2016multimodal} collected skeletal gesture expressions that frequently occur in game scenarios. Similarly, the emoFBVP \cite{ranganathan2016multimodal} dataset features a multi-modal actor recording that includes body gestures with skeletal tracking. The Emilya \cite{fourati2014emilya} dataset captures 3D body movements of posed emotions through a motion capture system.

\begin{table*}[htp]
  \centering
  \setlength{\abovecaptionskip}{0pt}
      \caption{The attributes comparison of iMiGUE with other widely used datasets for recognizing gesture-based expression of emotion. F/M: Female/Male, C: Controlled (in-the-lab), U: Uncontrolled (in-the-wild), SP: Spontaneous, F: Face,  G: Gesture, V: Voice, IMG: Identity-free Micro-Gesture.}
  \footnotesize
    \begin{tabular}{|l|c|c|c|c|c|c|c|c|c|c|c|}
    \hline
     \tabincell{c}{Datasets} & \tabincell{c}{\texttt{\#} Ge-\\stures} & \tabincell{c}{\texttt{\#} Em-\\otions}  &  \tabincell{c}{\texttt{\#}Subjects\\(F/M)}  & \tabincell{c}{\texttt{\#} Sam-\\ples} & \tabincell{c}{\texttt{\#}Vid-\\eos} & \tabincell{c}{Duration\\(sec)}  & \tabincell{c}{Con-\\text} & \tabincell{c}{Expr-\\ession} & \tabincell{c}{Resolution} & \tabincell{c}{Modalities} & \tabincell{c}{Recog-\\nition}\\
    \hline \hline
FABO \cite{gunes2006bimodal}                  & - & 10 & 23  (12/11)  & 206   & 23   & 360      & C  & Posed & 1024$\times$768   &F + G  &Isolated\\
HUMAINE \cite{douglas2007humaine}             & 8 &  8 & 10  (4/6)    & 240   & 240  & 5-180  & C  & Posed & -                &F + G  &Isolated\\
GEMEP \cite{glowinski2008technique}           & - & 18 & 10 (5/5)  & 7\,000+  & 1260 & -        & C  & Posed & 720$\times$576    &F + G  &Isolated\\
THEATER \cite{kipp2009gesture}                & - & 8  & -            & 258   & -    & -        & U  & SP  & -                &G  &Isolated\\
EMILYA \cite{fourati2014emilya}               & 7 & 8  & 11 (6/5)   & 7\,084  & 23   & 5.5    & C  & Posed &1280$\times$720   &G  &Isolated\\
\tabincell{l}{LIRIS-\\ACCEDE} \cite{gavrilescu2015recognizing} & 6 & 6  & 64  (32/32)  & -     & -    & 60      & C  & Posed & -                 &F + G  &Isolated\\
emoFBVP \cite{ranganathan2016multimodal}      & 23 & 23  & 10 (-)  & 1\,380  & -    & 20-66   & C  & Posed &640$\times$480   &F + G + V  &Isolated\\
BoLD \cite{luo2020arbee}                      & - & 26 & -    & 13\,239  & 9\,876    & -        & U  & SP  & -   &G  &Isolated\\
    \hline
iMiGUE (Ours)                                 & 32 & 2 & 72 (36/36)   & 18\,499 & 359 & \tabincell{c}{0.5-25.8 \\ (Min)} & U  & SP   & 1280$\times$720 & IMG     & Holistic\\
    \hline
    \end{tabular}
\label{tb:datasetscom}
\end{table*}%

Subsequent studies have focused more on studying spontaneous emotional gestures, which are more challenging compared to posed gestures. Video clips of emotional gestures extracted from two movies constitute the Theater dataset \cite{kipp2009gesture}, making the dataset very close to real-world scenarios. In \cite{kleinsmith2011automatic}, gestures while trying out a body movement-based video game are recorded to capture gestural movements. BoLD \cite{luo2020arbee} is a large-scale body expression dataset collected by Luo \textit{et al.}, which segments field-perceived emotional data from movies and reality TV shows. Until today, there have been few systematic studies of such fine-grained micro visuals of the body for MG, an essential cue for understanding repressed or hidden emotions.

\subsection{Methods for Gesture/Action Recognition.}
Early work on automatic modeling of emotional gestures relied heavily on hand-crafted features \cite{gunes2006bimodal,schindler2008recognizing,kleinsmith2011automatic}. Many recent works have introduced neural networks to accomplish gesture/action recognition. Supervised learning-based approaches have had great success in the past few years, giving birth to various works. The first attempts were based on two-dimensional (2D) convolutional neural networks (CNN) \cite{simonyan2014two, feichtenhofer2016convolutional,wang2018temporal,zhou2018temporal,lin2019tsm,yu2020searching}, which extract spatial features from the selected frames and implement temporal aggregation by fusing additional optical flow features or temporal pooling layers. There are many approaches that design filters in 3D to jointly capture temporal semantics, e.g., a 3D CNN \cite{tran2015learning,carreira2017quo,wang2018non,tran2018closer,hara2018can,xie2018rethinking,tran2019video}, which can be layered throughout the network to process temporal information. Further, some approaches complete the joint implementation of 3D CNNs and two-stream networks, such as the Slow-Fast approach \cite{feichtenhofer2019slowfast}. Unlike RGB-based methods, researchers increasingly favor skeleton data, benefiting from the properties of being lighter, faster processing, more robust to illumination, background, and actor appearance, and providing better protection for private biological information such as face information. Current work on skeleton-based action recognition can be divided into two main categories: RNN-based approaches \cite{du2015hierarchical,liu2016spatio,song2017end,liu2017global,zhang2017view,li2018independently}, which process the gesture skeleton directly as a time series, and approaches based on graph convolutional networks (GCN) \cite{yan2018spatial,li2018spatio,si2019attention,shi2019two,peng2020learning,cheng2020skeleton, liu2020disentangling}, which restructures the skeleton data into a graph.

Unsupervised-based action recognition is more challenging than supervised learning, and only a few attempts based on skeleton data have been reported. Zheng \textit{et al.} \cite{zheng2018unsupervised} proposed the LongT GAN (generative adversarial network) based on an encoder-decoder structure where the encoder learns the representation of the action and the features of the final state are used for the final action recognition task. The decoder regenerates the input sequence and the discriminator distinguishes whether the generated regeneration is correct or not. Su \textit{et al.} \cite{su2020predict} proposed a recurrent neural network based on encoder-decoder structure. The encoder learns separable representations to perform the prediction task. The decoder and encoder cluster similar motions into the same cluster and different motions into distant clusters in the feature space. However, these methods ignore the semantic association between encoder and decoder. Gao \textit{et al.} \cite{xuehaogao0contrastive} proposed an unsupervised action recognition method based on contrastive learning with ResNet as the encoder, using a generalized contrastive learning framework. Two skeleton augmentation methods, viewpoint and distance augmentation, are proposed. Rao \textit{et al.} \cite{Rao2021AugmentedSB} proposed an unsupervised action recognition method based on contrastive learning with momentum long short-term memory (LSTM). Momentum LSTM is used as the encoder for extracting representations and MOCO as the framework for contrast learning. Li \textit{et al.} \cite{li2021crossclr} proposed SkeletonCLR with ST-GCN as the backbone for skeleton contrastive learning, and CrosSCLR with information from other modalities to assist in mining positive/negative samples. 
Thoker \textit{et al.} \cite{10.1145/3474085.3475307} proposed inter skeleton contrastive learning, with streams of other input patterns to aid in the development of more comprehensive positive/negative samples. Lin \textit{et al.} \cite{10.1145/3394171.3413548} proposed MS$^2$L to learn more general representations by introducing multiple self-supervised tasks. Guo \textit{et al.} \cite{guo2022aimclr} proposed AimCLR to mine positive samples with extreme enhancement and energy-based attention-guided descent modules to improve the generalizability of the learning representation. Zhang \textit{et al.} \cite{lin2023actionlet} propose a hierarchical consistent contrastive learning framework (HiCLR) for skeleton-based action recognition, using a gradual growing augmentation policy and an asymmetric loss to enforce hierarchical consistency, demonstrating effectiveness through strong augmentation of 3D skeletons.
\section{Rethinking Micro-gestures}
\subsection{iMiGUE Dataset}
\begin{figure*}[htp]
\begin{center}
\includegraphics[width=16cm]{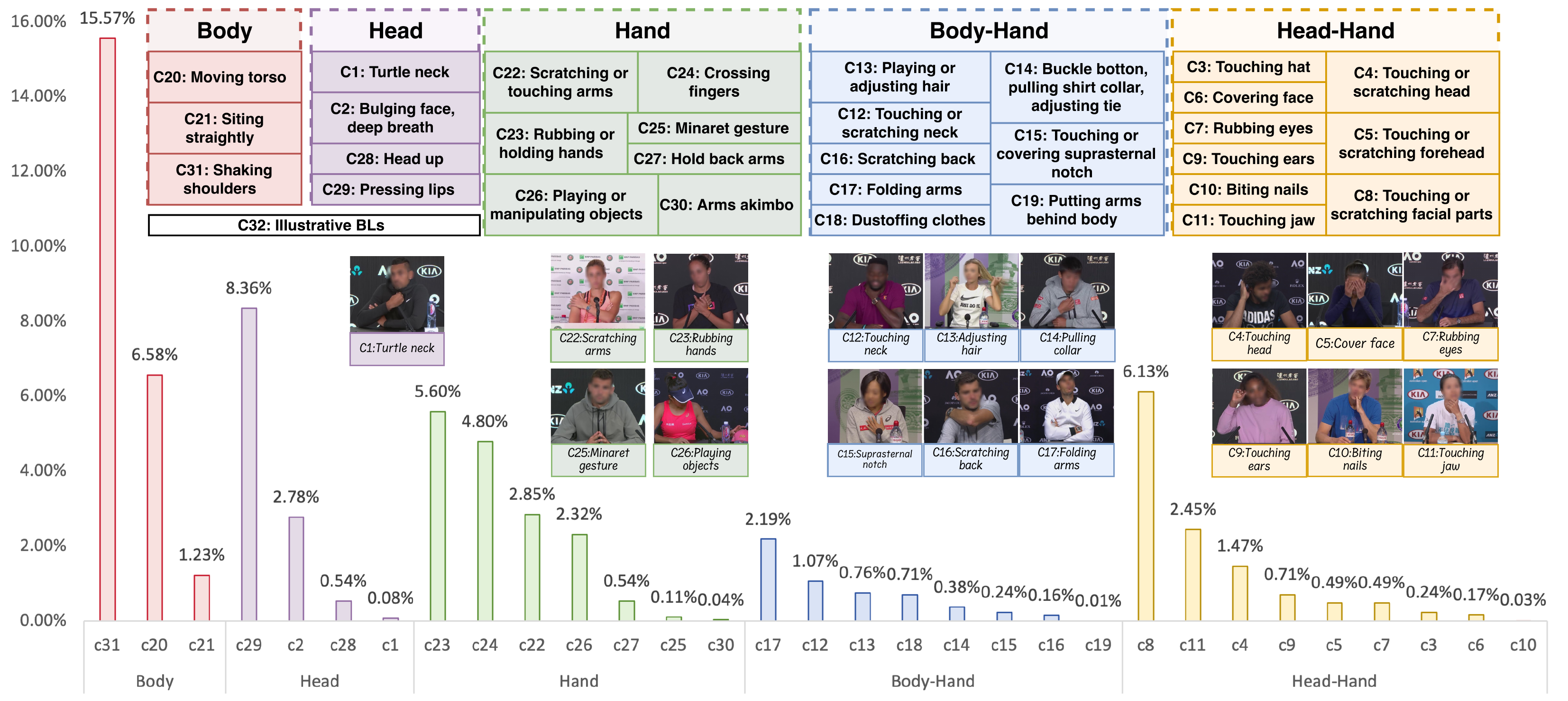}
\end{center}
\caption{Categorie distribution and examples of the iMiGUE dataset: Categories of MGs in iMiGUE dataset which refers to psychological studies \cite{ekman2009telling,pease2008definitive,navarro2016every}; Sample percentages of each category in the iMiGUE dataset; Examples (face masked) of MG categories in the iMiGUE dataset.}
\vspace{-0.4cm}
\label{fig:imigue}
\end{figure*}
Because iMiGUE is very different from previous gesture datasets, our work faces 
the following unprecedented challenges: 
i) Category definition. We refer to psychological researches\cite{ekman2009telling,pease2008definitive,navarro2016every} to more accurately define and organize the categories of emotion-related MGs. The focus is to distinguish the boundaries between MGs and illustrative behaviors clearly. ii) Elicitation and collection. We want the micro-gestures in the iMiGUE dataset to be actual occurrences, implying real emotions rather than posed gestures. Therefore, it is good to try to select and collect from real-world videos containing MG occasions, for example, from online video sharing platforms. Moreover, we want iMiGUE to go beyond identifying gestures and further explore the emotional states behind MGs, so finding and providing the underlying causes of these (MGs) emotional outbursts is also an essential task for us. iii) Data annotation. The quality of the labels 
directly affects the results of model training and testing. The iMiGUE dataset requires two levels of data annotation: labels for all MG occasions (clip-level) and labels for the corresponding emotion (video-level). We determine the MG labels based on the criteria of relevant psychological studies. Based on objective facts, ``positive'' and ``negative'' were used as the two emotional categories, corresponding to two scenarios of victory (positive emotion) and defeat (negative emotion) of the game. Due to MGs' complexity and diversity, we hired a team explicitly trained for the annotation process. We have also designed an effective quality control mechanism to ensure further high-quality annotations (see Sec. \ref{sec:datacon}).

\subsection{Dataset Construction}
\label{sec:datacon}
\subsubsection{Data collection}
Based on the above considerations, a series of videos (see Fig. \ref{fig:imigue}) using the ``post-match press conferences'' as a realistic scenario has been collected and organized. Because these press conferences were held immediately after the match, there was no (or negligible) additional time for the athletes to prepare for the reports' questions. In other words, even a seasoned athlete who can respond to these questions with ``witty'' statements cannot contain the unconscious leaks of MG. Because ``the subconscious mind acts automatically and independently of our verbal lie'' \cite{pease2008definitive}. The game's outcome will directly affect the athletes' positive or negative emotional state. Therefore, to study emotional AI, machines should be allowed to recognize and understand these identity-free MGs, and a good AI should be able to use them holistically to understand the players' emotional state.

The first data source for iMiGUE uses post-match press conferences from ``Grand Slam'' (tennis) tournaments because of several apparent advantages: i) High-quality data, at least 720p resolution. ii) Clean backgrounds, with no background distractions for the MG-centric instances. This is different from previous datasets where the backgrounds were either intrusive or could be distinguished between different categories. iii) Diverse cultures and nationalities. Tournament players come from almost every corner of the planet (see statistics of iMiGUE in Sec.\ref{sec:datastat}). iv) Good gender balance. There are 128 male and 128 female players in each Grand Slam, we can easily create a gender-balanced dataset.

\subsubsection{Data Annotation}
The iMiGUE annotates the included data at two levels: the MGs and the emotion categories. The dataset includes positive and negative emotional categories corresponding to winning and losing cases. Then, we searched for all MG instances (clips) in the videos and assigned corresponding category labels to them. Due to the difficulty and time-consuming nature of MG annotation, the following three measures were employed to ensure the annotation quality:
i) Clarify the scope and categories of MGs.
The referenced psychological studies \cite{ekman2009telling,pease2008definitive,navarro2016every} defined MGs into five major categories based on the location and function of the action, \textit{i}.\textit{e}., ``Head'', ``Body'', ``Hand'', ``Body-Hand'', and ``Head-Hand'', and each of which can be subdivided into more categories of MGs. A total of 31 categories of MGs and an additional category of non-MGs, \textit{i}.\textit{e}., illustrative gestures, are included in iMiGUE (see Fig. \ref{fig:imigue} for details).
ii) Multiple labelers and training for labeling.
The annotation of MG was done by five people together, based on two considerations: first, to speed up the process, and second, to reduce personal bias to obtain a more reliable annotation. All five people were trained to align their standards for MG annotation prior to the actual annotation.
iii) Cross check for reliable annotations.
Five annotators were responsible for annotating all video clips, and it was ensured that each clip had two annotators. After the annotation was completed, the following equation was used to cross-check the annotations: 
\begin{equation}\label{eq:reliab}
\mathcal{R} = \frac{2\times MG(L_{i},L_{j})}{\# All\_MG},
\end{equation}
where the number of MGs 
is 
$MG(L_{i},L_{j})$ that agreed by Labeler $i$ and Labeler $j$, the $\# All\_MG$ denoted the total number of MGs annotated by both labelers. 

The average inter-labeler reliability $\mathcal{R}_{avg}$ of iMiGUE is 0.81 which indicates reliable annotations. For the inconsistent annotation cases, the five labelers discussed them through and kept those with unified opinions while the rest (still with diverse opinions) were left out of the final label list.

\subsection{Dataset Statistics and Properties}
\label{sec:datastat}

In total, iMiGUE collected 359 post-match press conferences (258 wins and 101 losses) from Grand Slam tournaments, all sourced from online video-sharing platforms (\textit{e}.\textit{g}., YouTube), with a combined video length of 2\,092 minutes. Each video has a resolution of 1280$\times$720 pixels and a frame rate of 25 frames per second. A total of 18\,499 MG samples were labeled in these videos, with an average of about 51 MG samples per video, and all MGs are assigned 32 category labels. MG instances also have different lengths and an average length of 2.55\,s. iMiGUE's main feature figures compared to other gesture datasets are shown in Table \ref{tb:datasetscom}. It is worth noting that many spontaneous emotion datasets \cite{yan2014casme,li2013spontaneous,davison2016samm} have a common situation where the number of samples for different emotions is not balanced. The iMiGUE dataset also has this same feature with a large difference in the number of samples for the 32 MG categories (see Fig. \ref{fig:imigue}). This is a challenge for MG identification, which we will describe in more detail later.

Distinct from existing work, the iMiGUE dataset has 
very attractive features. i) {Micro-gesture-based dataset.} To the best of our knowledge, iMiGUE is the first publicly available micro-gestures dataset. ii) {Identity-free.} Sensitive biometric data such as face and voice have been masked and removed. iii) {Ethnic diversity.} iMiGUE data embraces and analyzes a wide range of cultures and contains 72 players from 28 countries and regions. iv) Gender-balanced. The iMiGUE dataset contains 36 female and 36 male players, ages ranging from 17 to 38 years. v) Winning and losing as the natural and objective reference for emotion categories. As a new dataset with many undetermined factors present, the results of the matches can be used as a more objective reference for emotional states. We suggest that researchers who work on estimating emotional states but are concerned about privacy issues can use this dataset as a benchmark.
\section{MG recognition and Emotion Understanding}
\label{sec:MGandEMO}
\subsection{Rethinking Micro-Gesture Recognition}
Once the dataset is constructed, a crucial issue that requires thorough consideration is the problem of unbalanced data, which is often inevitable in an in-the-wild setting, unlike controlled recording situations where sample sizes are fixed or planned. The iMiGUE dataset, being an in-the-wild environment dataset, is inherently susceptible to class imbalance issues characterized by a long-tailed distribution (refer to Fig.~\ref{fig:imigue}). Such a scenario could potentially present significant difficulties for fully-supervised learning models, leading to a considerable reduction in performance in the presence of extreme label bias. Unsupervised models offer an intuitive alternative to fully supervised methods as they do not require human-labeled data. In our previous work \cite{Liu_2021_CVPR}, we proposed the U-S-VAE, an encoder-decoder model that maximizes mutual information between input and re-generated output using KL-divergence. However, the U-S-VAE has not fully explored the semantic association between the encoder and decoder, and contrastive learning has emerged as a superior self-supervised approach for representation extraction. Therefore, we designed a Spatial-Temporal-Balanced Dual-stream-contrastive learning network to improve MG recognition. Our method differs from previous contrastive learning models in two ways: i) micro-gesture augmentations, where we specifically explore new augmentations for MG characteristics; ii) spatial-temporal-balanced fusion, where we organize data into different forms, forcing the model to learn temporal and spatial features in different types of data and encoders.

\subsubsection{Skeletal Micro-Gesture Augmentations Exploration}
\label{sec:augmentation method}

Contrastive learning employs positive and negative pairs to achieve consistent representations for similar instances and distinct representations for dissimilar instances. The creation of positive and negative pairs is crucial to contrastive learning. Augmentations of an instance are usually considered positive pairs, whereas augmentations from different instances serve as negative pairs. Previous research \cite{Rao2021AugmentedSB} proposed augmentation techniques for action recognition based on skeletons. However, MGs have distinct characteristics from general actions, such as subtle movements. Therefore, some original augmentation techniques may not be suitable for MGs, and specific augmentation techniques must be designed for them.

MGs possess certain distinctive features, namely: i) a limited number of movement joints and small movement amplitudes, such as shrugging shoulders and sitting upright, among others, in the spatial domain; ii) brief duration and ease of segmentation into repetitive (or static) action segments, such as rubbing hands, rubbing eyes, and grabbing the head, among others, in the temporal domain; and iii) susceptibility to misclassification, as alterations in joint coordinates or changes in perspective can easily lead to their misidentification as other actions, such as touching the neck being misinterpreted as touching the face or adjusting hair. To exploit these unique attributes, we explore three types of augmentation strategies to extract positive samples pertinent to micro-gestures effectively.

The joint movements in MGs are often subtle and can easily be confused, leading to category ambiguity. As shown in Fig.~\ref{fig:imigue}, it displays several MGs, such as ``touching the neck," ``touching the suprasternal notch," ``touching the hat," ``touching the ear," ``covering the face," and ``touching the jaw." However, these MGs may transform into a different category after being perturbed. For instance, ``touching the neck" may become ``touching the jaw", while ``touching the hat" could become ``touching the ear." Due to their subtle nature, the differences between MGs can be minuscule. By introducing slight perturbations to MGs, we can generate challenging samples within the same category, making it more likely to distinguish between categories that are often confused. We can transform an MG into a confusing category with a low probability of learning the consistency between such categories. This plays a crucial role in exploring the boundaries between different categories of MGs. We strive for the encoder trained through the proposed model's contrastive learning process to withstand minor noise, bolstering the model's robustness and ensuring adaptability to individual differences among MGs in real-world scenarios. 
To achieve this goal, we explore augmentation strategies involving joint perturbation and view perturbation.

\begin{figure*}[!h]
\begin{center}
\includegraphics[width=17cm]{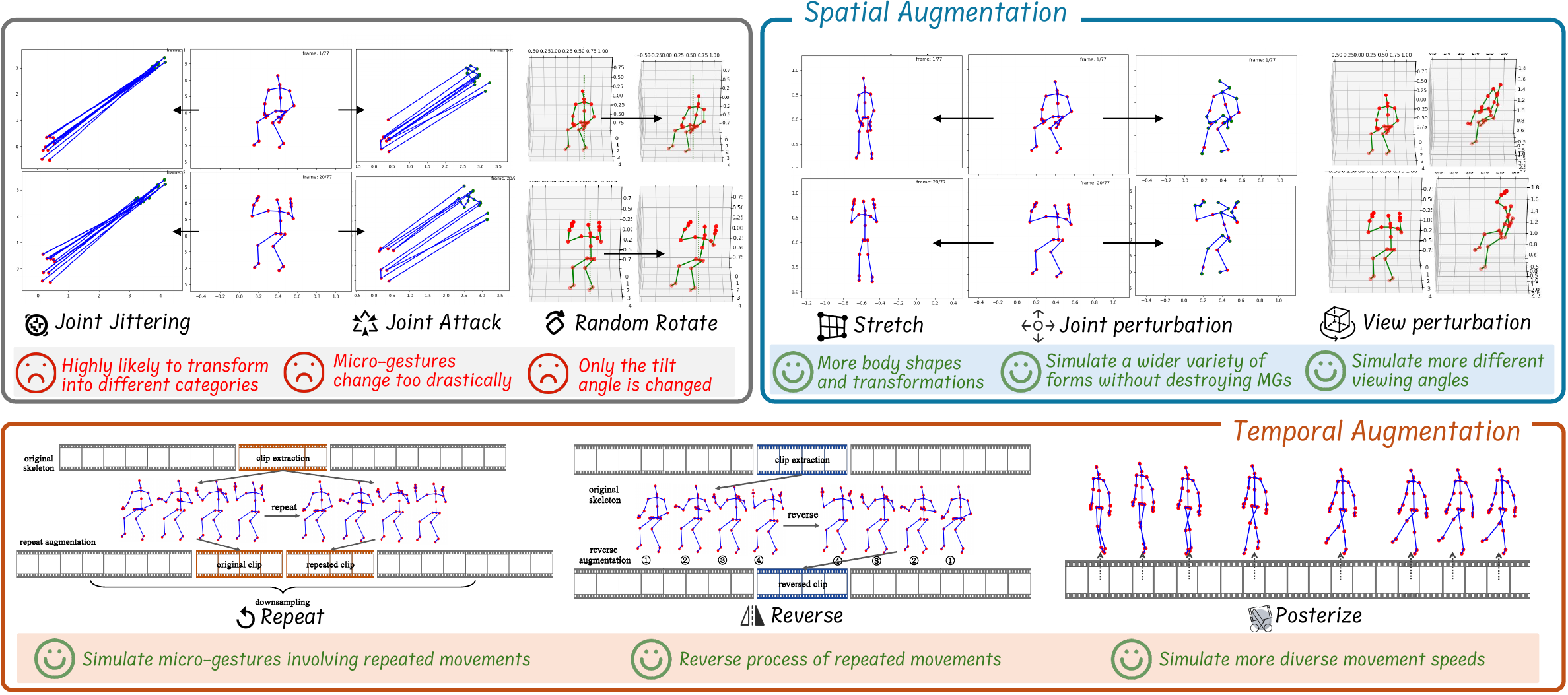}
\end{center}
\caption{Micro-gesture skeleton data augmentation methods.}
\vspace{-0.2cm}
\label{fig:MG_all}
\end{figure*}

\textbf{Coordinate perturbation.} Estimating the magnitude of noise added to a joint is challenging. Excessive joint jitter can lead to MGs deformation being classified as a completely different category. Such significant augmentations should be defined as attack-style augmentations, such as joint masking in AS-CAL\cite{Rao2021AugmentedSB} and joint shaking in ISC\cite{10.1145/3474085.3475307}. In joint shaking augmentation methods, joints move within the range of their coordinates multiplied by $[-1,1]$, which seems like a small interval, and these augmentation methods are effective for illustrative behavior recognition. However, this may cause MGs distortion that would never occur in natural environments. This creates a greater attack on MGs, which to some extent limits the ability to explore realistic and more challenging representations of motion that may arise.

For MGs, the following two cases may occur. The first is round-trip motion: the motion ends and almost resumes the original posture (e.g., shrugging shoulders, etc.), and the second is one-way motion: the motion ends and maintains the state (e.g., crossed fingers, etc.). We select the start frame $L_{sta}$, the middle frame $L_{mid}$, and the final frame $L_{fin}$ of the coordinates in all $T$ frames, and calculate the perturbation coefficient by the following equation, in which the $\tau$ is temperature coefficient:
\begin{equation}\label{eq:bj}
\lambda = \frac{1}{\tau} \frac{L_{fin} - L_{mid}}{L_{mid}-L_{sta}}.
\end{equation}

Using the original joint as the starting point and $R$ as the perturbation matrix, the original coordinates are perturbed to another position. 
\begin{equation}\label{eq:cj}
{A_{cp}}\left( {{x_n}} \right){\rm{ = }} x_n+ \lambda R,
\end{equation}
where $x$ is the joint coordinate matrix, $n$ is a randomly drawn joint ordinal number, and $x_n\in\mathbbm{R}^{1\times3}$ denote the coordinate matrix of the extracted node at a specific moment. A random perturbation matrix $R\in\mathbbm{R}^{1\times3}$ is created, with its elements drawn from a uniform distribution over $[-1,1]$. For each frame, $m$ joints are extracted and their coordinates are perturbed using different perturbation matrices, with the same perturbation matrix used for each joint across different time frames. As shown in Fig.~\ref{fig:MG_all}, our proposed ``joint perturbation" method provides more possibilities for the MGs while preserving the original spatial structure and natural semantics. The model is encouraged to prioritize temporal joint motion over relying solely on the skeletal structure connection, resulting in increased robustness to joint noise across different motion patterns.
\begin{figure*}[!h]
\begin{center}
\includegraphics[width=13cm]{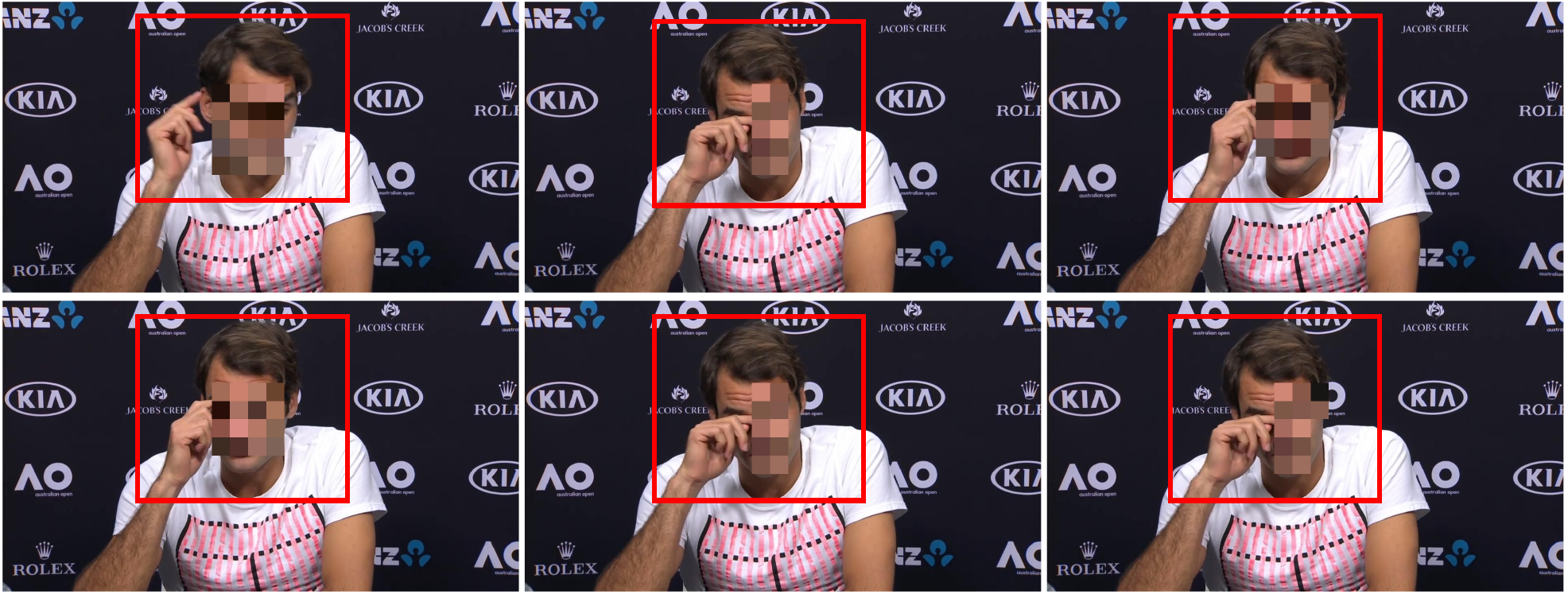}
\end{center}
\caption{Repeatability example of micro gestures.}
\label{fig:MG_1}
\end{figure*}

{\bf{View perturbation.}} Similar to the perturbation of coordinates, the trained model is also expected to resist the perturbation of view changes and to have consistent prediction results for observations from different angles. Inspired by \cite{Rao2021AugmentedSB} which employed a random rotation augmentation method but mainly focused on rotating the skeleton in one axis while leaving the other axes relatively unchanged. Observers in real life can observe both regular behaviors and MGs from any angle of observation. Therefore, we have developed a view perturbation augmentation method that can simulate changes in the observation angle. Euler angles are a set of three independent angle parameters, which are composed of nutation angle $\theta_x$, precession angle $\theta_y$, and rotation angle $\theta_z$, used to uniquely determine the position of a fixed point rotating rigid body. According to Euler's formula, we give the rotation arithmetic matrix in right-handed coordinates, on the X, Y, and Z axes, as follows:

\begin{equation}\label{eq:rx}
{R_X}\left( {{\theta _x}} \right){\rm{ = }}\left[ {\begin{array}{*{20}{c}}
1&0&0\\
0&{\cos {\theta _x}}&{\sin {\theta _x}}\\
0&{ - \sin {\theta _x}}&{\cos {\theta _x}}
\end{array}} \right],
\end{equation}

\begin{equation}\label{eq:ry}
{R_Y}\left( {{\theta _y}} \right){\rm{ = }}\left[ {\begin{array}{*{20}{c}}
{\cos {\theta _y}}&0&{ - \sin {\theta _y}}\\
0&1&0\\
{\sin {\theta _y}}&0&{\cos {\theta _y}}
\end{array}} \right],
\end{equation}

\begin{equation}\label{eq:rz}
{R_Z}\left( {{\theta _z}} \right){\rm{ = }}\left[ {\begin{array}{*{20}{c}}
{\cos {\theta _z}}&{\sin {\theta _z}}&0\\
{ - \sin {\theta _z}}&{\cos {\theta _z}}&0\\
0&0&1
\end{array}} \right].
\end{equation}

In the view perturbation augmentation, the changed view is obtained by rotating one dimension at $[-\frac{\pi}{4},\frac{\pi}{4}]$ and the remaining two dimensions at $[-\frac{\pi}{6},\frac{\pi}{6}]$, and the skeleton is regenerated at that angle for each frame to maintain the consistency of the MG. As shown in Fig.~\ref{fig:MG_all}, view perturbation augmentation is able to simulate more diverse viewpoint changes instead of being limited to rotation on one axis only.

MGs are actions that occur in the subconscious mind and are influenced by the actor's psychological and physiological states. As a result, these unconscious actions often involve repetitive (or static) movements and are influenced by an individual's habitual movements. Based on such characteristics, we propose utilizing clip reverse, clip repetition, and posterize time to enhance the model's ability to learn a robust representation of temporal changes.

{\bf{Repeat Clip.}} In the case of MGs that involve repetitive actions, such as rubbing the eyes (as shown in Fig.~\ref{fig:MG_1}), the frequency of hand rubbing does not affect the categorization of this MG. Conversely, for MGs that do not contain repetitive actions, like sitting up, repeating a single action introduces a certain amount of temporal noise to the MG, but it does not alter the overall motion trend. Augmenting the MG dataset with repeated clips can enhance the model's ability to extract features of the motion trend in the presence of temporal noise, and enhance its robustness to repetitive actions in the MGs.

The clip repetition augmentation is illustrated in Fig.~\ref{fig:MG_all}. In this approach, a skeleton sequence consisting of $T$ frames is considered. The starting frame $L_b$ of the repeated clip is randomly selected from the interval $[0, L_0]$, where $L_0$ represents a configurable starting point. Additionally, the length of the repeated clip, denoted as $L_d$, is randomly chosen from the interval $[L_{low}, L_{high}]$, with the constraint that $L_0 + L_{high} < T$. To generate a longer sequence, denoted as $\hat{L}$, the extracted clip is inserted into the sequence at the position $L_b$ and spans $L_d$ frames. Subsequently, $\hat{L}$ is down-sampled to match the original sequence length, preserving the overall trend of the MG.

{\bf{Reverse Clip.}} As illustrated in Fig.~\ref{fig:MG_all}, the repeat clip represents the forward motion of an MG, while the reverse clip corresponds to its reverse motion. It is worth noting that the repetition process of the clip necessitates a bidirectional operation, and dividing it into unidirectional components would amplify the semantic complexity of the MG. Firstly, a clip of length $L_v$ is chosen, and subsequently, its sequence is inverted and reintegrated into its original position, generating a new sequence. It is compelled to acquire a more coherent representation by subjecting the model to this temporal interference.

{\bf{Posterize Time.}} The temporal allocation of movement varies from person to person and is influenced by individual movement habits. The ``posterize'' in Fig.~\ref{fig:MG_all} illustrates the process of posterize time augmentation. Initially, two small clips are sampled at the beginning and end of the sequence to ensure the overall trend of MGs remains consistent. Subsequently, several clips are randomly selected at different interval frequencies, effectively slowing down certain parts while accelerating others. This transformation effectively captures the action variability of MGs across individuals. Through contrastive learning, the objective is to learn the shared characteristics of similar MGs across different individuals within these variations.

To enhance the efficacy of contrastive learning in generating more cohesive representations within positive samples within a lower-dimensional abstract potential feature space, it is imperative to devise advanced augmentation techniques that promote implicit consistency within a higher-dimensional concrete real space. Additionally, it is essential to ensure that our feature encoding accounts for more discriminative spatial semantics. Given the significant variations in body size and action patterns among different actors, we propose the incorporation of a novel augmentation called ``stretch."

{\bf{Stretch.}}
We define two skeletal variation matrices for body shape and tilt (inspired by \cite{Rao2021AugmentedSB}) as follows:

\begin{equation}\label{eq:shape}
{A_{shape}}\left( {{X}} \right){\rm{ = }}X\cdot{\left[ {\begin{array}{*{20}{c}}
\alpha&0&0\\
0&\beta&0\\
0&0&\gamma
\end{array}} \right]},
\end{equation}

\begin{equation}\label{eq:tilt}
{A_{tilt}}\left( {{X}} \right){\rm{ = }}X\cdot{\left[ {\begin{array}{*{20}{c}}
1&t_x^y&t_x^z\\
t_y^x&1&t_y^z\\
t_z^x&t_z^y&1
\end{array}} \right]},
\end{equation}
where $\alpha$, $\beta$, $\gamma$ are randomly sampled form $\left[1,2\right]$ and $t_x^y$,$t_x^z$,$t_y^x$,$t_y^z$,$t_z^x$,$t_z^y$ are randomly sampled from $\left[-1,1\right]$. For a micro-gesture sequence, we randomly apply one transformation matrix to it.

As depicted in Fig.~\ref{fig:MG_all}, the incorporation of this strategy allows for the simulation of a wider array of MGs, accommodating various body types. Consequently, this augmentation technique significantly enhances the diversity of the samples, thus augmenting the richness of the dataset.

\subsection{Spatial-Temporal-Balanced MG Recognition Method}
\label{sec:contrastive method}
For Spatial-Temporal-Balanced Dual-stream Fusion, spatial and temporal streams occupy an equally important position. Since they are both built based on a contrastive learning structure, their processes have a certain degree of similarity. In this subsection, we present how to build a single-stream MG recognition model based on contrastive learning and adaptive graphs, using GCN-based spatial streams as an example. Before delving into our proposed architecture, we first briefly introduce some key details of the adaptive graph.

\subsubsection{Preliminary}
A human skeleton sequence is denoted as $\mathcal{G}=(\mathcal{G}_1,...,\mathcal{G}_T)$ and contains $T$ consecutive frames, where $\mathcal{G}_i$ is denote the ${i^{th}}$ frame human skeleton graph. For each graph $\mathcal{G}=(\mathcal{V},\mathcal{E})$, where $\mathcal{V}=\{v1,...,v_n\}$ is a set of N nodes that represent joints, $\mathcal{E}$ is a set of edges that represent bones, which is denoted as adjacency matrix $A\in\mathbb{R}^{N\times N} $. The element $a_{i,j}=1$ and it symbolizes whether joint $v_i$ and joint $v_j$ are connected by bone. If it is true $a_{i,j}=1$ also $a_{j,i}=1$ since $\mathcal{G}$ is an undirected graph, otherwise $a_{i,j}=0$ and $a_{j,i}=0$. The feature matrix $X \in \mathbb{R}^{T\times N\times C}$ represents node features set $\mathcal{X}$ for the skeleton sequence, and $X_t\in \mathbb{R}^{N\times C}$ is the node feature at time t. $\Theta^{(l)}\in \mathbb{R}^{C_l\times C_{l+1}}$ is a learnable weight matrix at layer $l$ of a network.

\begin{figure*}[t]
\begin{center}
\includegraphics[width=16cm]{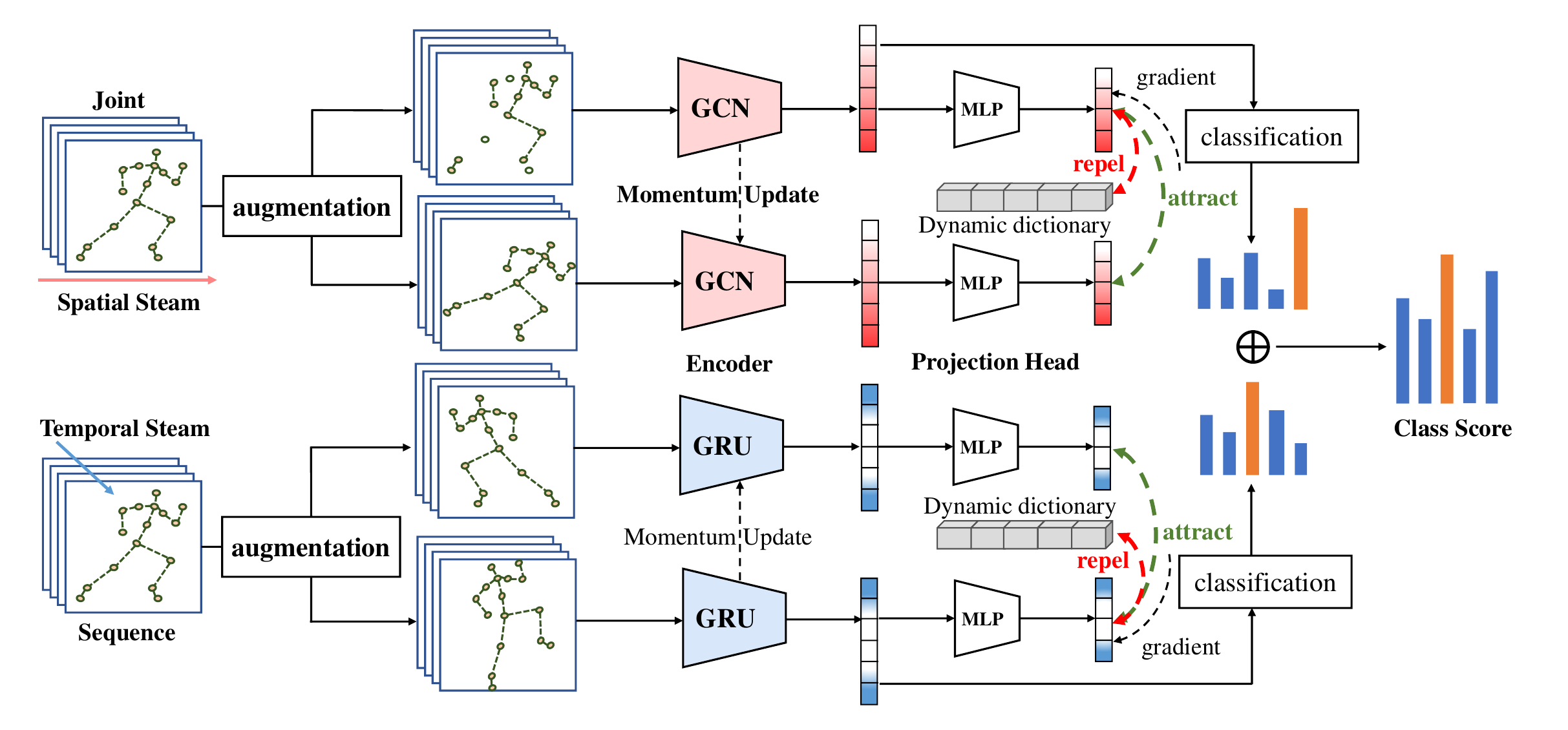}
\end{center}
\vspace{-0.45cm}
\caption{Spatial-Temporal balanced dual stream contrastive learning framework.}
\vspace{-0.4cm}
\label{fig:ST-CLR}
\end{figure*}

For given graph $\mathcal{G}=(\mathcal{V},\mathcal{E})$, the inputs to GCN are its feature matrix $X$ and adjacency matrix $A$. Denote $\tilde{A}=A+I$, where $I$ is the identity matrix meaning self-loops. $\tilde{D}$ is the diagonal degree matrix of  $\tilde{A}$. The features at time $t$ can be derived from the GCN layer-wise update rule shown as Eq. \ref{eq:gcn}.
\begin{equation}\label{eq:gcn}
X_{t}^{l+1} = \sigma(\tilde{D}^{-\frac{1}{2}}\tilde{A}\tilde{D}^{-\frac{1}{2}}X_{t}^{l}\Theta^{(l)}),
\end{equation}
where $\sigma(\cdot)$ is an activation function. Eq. \ref{eq:gcn} shows that the information of the next layer of each node is obtained by weighted aggregation of the information of the previous layer itself as well as the neighboring nodes. Then, through the linear transformation $\Theta^{(l)}$ and the nonlinear transformation $\sigma(\cdot)$.

GCN-based methods contain two parts: spatial graph convolution (SGC) and temporal graph convolution (TGC). For temporal graph convolution, we use 1D convolution as TGC.

We follow the previous method and classify the joints into three classes $P=\{centripetal, root, centrifugal\}$ to obtain the GCN update rule as Eq. \ref{eq:gcn_new}:
\begin{equation}
\label{eq:gcn_new}
X_{t}^{(l+1)} = \displaystyle\sum_{p\in P}\sigma(\tilde{D}_p^{-\frac{1}{2}}\tilde{A}_p\tilde{D}_p^{-\frac{1}{2}}X_{t}^{(l)}\Theta_p^{(l)}).
\end{equation}

Given a mask matrix $M$, the original adjacency matrix $A$ determines whether there is a connection between two vertices, and $M_k$ determines the strength of the connection. To make the topology of the graph adaptive, the above equation is converted as Eq. \ref{eq:gcn_bk}:

\begin{equation}
\label{eq:gcn_bk}
X_{t}^{(l+1)} = \displaystyle\sum_{p\in P}\sigma(\tilde{D}_p^{-\frac{1}{2}}(B_p+\alpha C_p)\tilde{D}_p^{-\frac{1}{2}}X_{t}^{(l)}\Theta_p^{(l)}).
\end{equation}

The adjacency matrix of the graph is divided into two subgraphs: $B$ and $C$. $B$ is a global graph learned from the data, and $C$ is an individual graph that learns a unique topology for each sample. The parameter $\alpha$ is unique for each layer and is learned and updated during the training process.

In addition, the spatial attention module, temporal attention module, and channel attention module are used to help the model give different levels of attention to different joints, frames, and channels.

\subsubsection{Adaptive Graph Contrastive Learning}

Previously proposed GCN-based skeleton action recognition methods learn better representations by minimizing the loss function between the prediction result and ground truth which are manually labeled. In contrastive learning, the network optimizes the representation step by step by minimizing the similarity between positive pairs. The positive pairs are derived from the diverse augmented samples constructed in the augmentation block, and the representations are learned by the GCN encoder. However, directly using this approach may lead to two fatal problems: i) the information contained in the representation is unstable, and important personalized fine-grained features are lost when advancing the distance between positive sample pairs by similarity. ii) The network collapses, outputting the same features for all inputs. To cope with the first problem, we re-project the representation extracted from the adaptive GCN encoder to a latent space. To cope with the second problem, we adopt the approach of momentum updating network parameters and storing a large number of negative samples through dictionaries during contrastive learning. 

In the following, we will present in-depth the various parts of the framework for adaptive graph-based contrastive learning.

{\bf{Reprojection adaptive GCN-based encoder.}} For feature matrix $X$ extracted from the GCN, it is fed into a reprojection network consisting of MLP, ReLU, and MLP through project representation into a latent space to dig deeper and more personalized features. The reprojected feature $Z$ can be calculated by Eq. \ref{eq:reproject}:
\begin{equation}
\label{eq:reproject}
\begin{split}
Z_{t} = w_2\sigma(& w_1\sum_{p\in P}\sigma(\tilde{D}_p^{-\frac{1}{2}}(B_p+\alpha C_p) \\
                  & \tilde{D}_p^{-\frac{1}{2}}X_{t}^{(L-1)}\Theta_p^{(L-1)})+b_1)+b_2,
\end{split}
\end{equation}
where $w_1$ and $w_2$ are connection weights, $b_1$ and $b_2$ are connection biases.

{\bf{Dynamic dictionary.}} The high performance of contrastive learning is inextricably linked to a sufficient number of negative pairs. To provide the model with sufficient negative pairs, we introduce a dynamic dictionary to the contrastive learning block. The construction of a dynamic dictionary requires ensuring the following two conditions: 1) the dictionary needs to be large enough to contain enough negative pairs, which facilitates the extraction of good features. 2) the encoder parameters of the dictionary should be updated with momentum to maintain consistency. To satisfy these two conditions, the dictionary is treated as a queue, by which the size of the dictionary is made decoupled from batch size. Each epoch sends samples of batch size into the dictionary, and when the queue is full, the oldest batches in the queue are eliminated.

{\bf{Momentum update parameters.}} Using a queue expands the size of the dictionary, but it becomes harder to back-propagate to the key encoder since the gradient is propagated across all the data in the queue. Simply copying the query encoder parameters directly to the key encoder breaks the consistency of the key-value representation. Therefore, the method of momentum update for parameters is used for the key encoder as seen in Eq. \ref{eq:update}.
\begin{equation}
\label{eq:update}
\xi\gets m\xi+(1-m)\theta,
\end{equation}
where $\theta$ is the parameters of the query encoder and $\xi$ is the parameters of the key encoder. This makes the key encoder slowly converge to the query encoder, but never reach the same parameters. Thus, it is possible to accumulate features of actions belonging to the same class by different augmentations and continuously optimize the model. This slow momentum update motivates the model to learn the features of each type of action under various augmentations.

{\bf{Contrastive loss.}} Obviously, in the process of contrastive learning, we hope to learn the inner consistency of the micro gestures implied in the diverse external expressions. It is hoped that the samples belonging to the same category will be pushed closer together and those not belonging to the same category will be distanced in the latent space - that is, to maximize the similarity between augmented instances. For query and key queue, query and its augmented samples are positive pairs, thus, the similarity between them should be increased. Query and other augmented samples are negative pairs, and the similarity between them should be reduced. We use the Eq. \ref{eq:pos} to calculate the angle $\theta^+$ between vectors $z_\theta$ and $z_\xi^+$:

\begin{equation}
\label{eq:pos}
\theta^+ = \frac{<z_\theta,z_\xi^+>}{\|z_\theta \|_2\cdot \|z_\xi^+\|_2}.
\end{equation}

Similarly, we use Eq. \ref{eq:neg} to calculate the angle $\theta^-$ between $z_\theta$ and $z_\xi^-$:

\begin{equation}
\label{eq:neg}
\theta^- = \frac{<z_\theta,z_\xi^->}{\|z_\theta \|_2\cdot \|z_\xi^-\|_2}.
\end{equation}

The smaller the angle $\theta^+$, the closer the distance of positive pairs. The larger the angle $\theta^-$, the farther the distance of negative pairs. We use the dot product to measure these angles and InfoNCE to calculate contrastive loss as Eq. \ref{eq:contrastive_loss}:

\begin{equation}
\label{eq:contrastive_loss}
\mathcal{L}_{i,j} =-\log \frac{\exp({\rm sim}(z_i,z_j)/\tau)}{\sum_{k=1}^K\mathbb{1}_{[k\neq i]}\exp({\rm sim}(z_i,z_k))/\tau}. 
\end{equation}

We use the following Eq. \ref{eq:sim} to calculate ${\rm sim}(z_i,z_j)$, which represents the dot product between $\l_2$ normalized $z_i$ and $z_j$, $\mathbb{1}_{[k\neq i]}$ evaluating to 1 if $k\neq i$, and $\tau$ is a temperature parameter, as:

\begin{equation}
\label{eq:sim}
{\rm sim}(z_i,z_j)=\frac{z_i\cdot z_j}{\|z_i \|_2\cdot \|z_j\|_2}.
\end{equation}

\subsubsection{GRU-based Contrastive Learning}
\label{sec:gru-CLR}

The GRU-based temporal stream contrastive learning network is built with the same steps as the above GCN-based spatial stream network. The difference lies in the different types of encoders used in contrastive learning. In addition, we emphasize that the projection head that mapped the features extracted from the encoder to the latent space is also different from that in the spatial stream, and the weights of the projection head will be learned during the training process. 

\subsubsection{Spatial-Temporal Fusion}
\label{sec:dual}

Many of the previous GCN-based supervised skeleton action recognition works \cite{yan2018spatial,li2018spatio,si2019attention,shi2019two,peng2020learning,cheng2020skeleton, liu2020disentangling} have used a multi-stream fusion strategy to motivate the model to integrate multiple information and improve the comprehensiveness of the perception of the action. However, since their networks are based on GCN, the input data are organized in the form of graphs. In general, there are three data input forms: joint, bone, and motion. The joint is the most basic node data, which represents the coordinates of each joint; the bone is calculated based on joint data according to the human skeleton structure, which represents the length and direction information of the bone; the motion is based on the difference between two adjacent frames also calculated from the joint data, which indicates the motion information of the joint. Many unsupervised-based works continue this approach, using the same multi-stream fusion method to enhance the model. However, such fusion is actually ``spatial heavy'' and is not obvious for temporal information. The RNN is commonly used for temporal integration. For example long and short-term memory (LSTM) \cite{donahue2015long, du2015hierarchical, veeriah2015differential, zhu2016co,Mahasseni_regularized_lstm:2016, li2016online,liu2016spatio,liu2017global} based approaches have demonstrated their advantages in learning sequential data. 

With this in mind, we went beyond the constraints of ``same data input form'' and ``same network architecture'' and designed a new ``Temporal-Spatial-Balanced Dual-Stream Fusion'' network. The overall architecture is shown in Fig.~\ref{fig:ST-CLR}. The data is organized as graphs and input to the GCN-based contrastive learning network in the spatial stream. The data is organized as sequences and input to the GRU-based contrastive learning network in the temporal stream. The softmax scores are calculated separately for the obtained spatial and temporal features and then summed to obtain the fused prediction scores. This allows us to draw on the advantages of both the GCN network in capturing structured information of bones and the GRU network in capturing long-range temporal information. When performing MG recognition provides a more comprehensive understanding of a specific sample, this leads to more accurate results. 

The overall process of our method is shown in Fig.~\ref{fig:ST-CLR}. To summarize, the network flow is as follows: i) Two different random augmentation methods are adopted for the input skeleton sequences to obtain positive pairs. Our augmentation methods are divided into three types: spatial augmentation, temporal augmentation, and perturbation augmentation, which fully mined the hard positive pairs. In contrast, the hard positive pairs of an instance become hard negative pairs for other instances. ii) A dynamic key dictionary queue is constructed to store sufficient negative pairs to the input query for contrastive learning. When the queue is full, the oldest batches in the queue are dequeued and new samples are enqueued. iii) Reprojected features extracted by the spatial encoder and temporal encoder enable the network to learn deeper and more detailed features. iv) The parameters of the query encoder and the key encoder are not the same. Key encoder momentum updates its parameters, slowly leaning on the query encoder but never the same with it, thus keeping the query-key representation consistent. v) Constructing the contrast loss maximizes the similarity between positive pairs and minimizes the similarity between negative pairs. Thus, instances of the same category are closer together in latent space, while instances of different categories are further apart. vi) Finally, we feed the test sequences into the trained spatial and temporal encoders to extract the spatial and temporal features. Then, we feed the obtained features into the spatial and temporal classification headers, respectively, and sum up the obtained softmax scores to find the category with the highest score as the predicted category. This completes the temporal-spatial-balanced fusion.

\subsection{Rethinking MG-based Emotion Understanding}
In practical situations, micro-gestures often accompany other modalities of information, such as dialogue and facial expressions. Individuals with rich social experiences typically integrate these cues to infer the current emotional state of those they interact with, thereby adjusting their communication strategies. In other words, micro-gestures often play a supportive role in real-life scenarios, with judgments rarely relying solely on them. In this section, we propose ingenious methods to validate the existence of this supportive role. This not only solidifies researchers' understanding of the information conveyed through micro-gestures in the pathway of emotions but also indicates new directions for further exploration.
\begin{figure*}[t]
\begin{center}
\includegraphics[width=17cm]{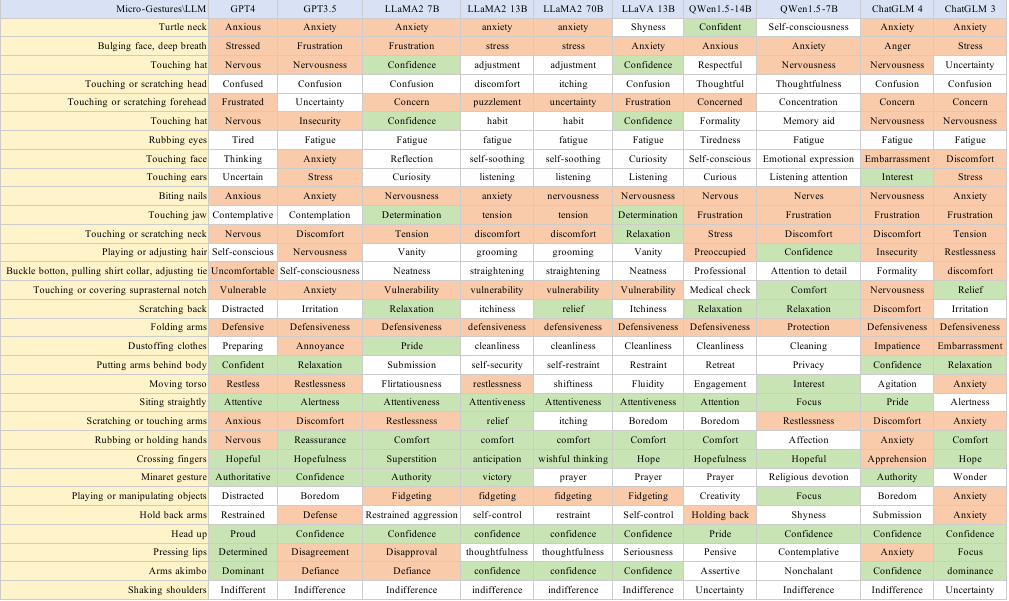}
\end{center}
\caption{Large language models' understanding evaluation of the relationship between micro-gestures and emotions. The orange background represents negative emotions, the green background represents positive emotions, and the white background represents neutral emotions.}
\label{fig:LLM_MGandEmo}
\end{figure*}
\subsubsection{Large language model prior knowledge evaluation}
\label{sec:LLM_Prior}
Considering the extensive knowledge boundaries of large language models, we have decided to use a large language model as the subject in this complex scenario. This model will act as a psychology expert with certain knowledge in psychology, tennis events, and an understanding of the relationship between psychology and micro-gestures to some extent. However, before we commence, considering that existing large language models are mostly trained on extensive datasets, to ascertain these models' understanding of micro-gestures and the relevant information in the iMiGUE dataset, we have devised a small evaluation bench: iMiGUE-Bench, to assess the prior knowledge of large models from two dimensions.

{\bf{MG and Emotion Knowledge.}} To validate whether large language models understand micro-gestures and their connection to emotions, we evaluated the understanding of 10 commonly used large language models regarding micro-gestures. The prompt format used was: "[What emotion do you think the \{Micro-Gesture\} below represents?]" Fig.~\ref{fig:LLM_MGandEmo} illustrates the emotions these large models perceive as most relevant to each micro-gesture, with orange background indicating negative emotions, green background indicating positive emotions, and white background indicating neutral emotions. It can be observed that these LLMs not only demonstrate an understanding of the association between micro-gestures and underlying emotions but also achieve consensus in many cases. Most LLMs tend to associate micro-gestures with negative or neutral emotions. However, there are still many micro-gestures classified as positive by the majority of LLMs. This is crucial for our subsequent use of LLMs and micro-gestures to infer emotions, as the number of micro-gestures classified as positive is fewer than negative and neutral ones, but they play a crucial role in comprehensive judgments.

{\bf{Match Facts Knowledge.}} Due to the iMiGUE dataset containing many sports interview contents, and the emotions of athletes are directly related to the outcome of the game, in such cases, LLMs are likely to know the objective results of these events. Therefore, we conducted two different validations: first, whether LLMs directly understand the outcomes of these events. For the 480+ events contained in iMiGUE, we designed the match results as multiple-choice questions. LLMs were prompted in the following format: "[In the \{Man's/Women's Singles\} \{Number\}st Round of the \{Year\} \{Match Name\}, the match between \{Player 1\} and \{Player 2\}, who is the winner? A. \{Player 1\}; B. \{Player 2\}]." Most LLMs performed well in this assessment, indicating that they might have been fed relevant data, especially GPT-4 achieved over 90\% accuracy. Next, considering the format of post-match interviews in iMiGUE, we also verified whether LLMs could infer the objective results of the matches from the information in the dialogues. In this evaluation, most LLMs, except for GPT-3.5 and GPT-4, showed significant room for improvement, with GPT-4 achieving only 40\% accuracy (correctly inferring both the interviewee's name and the match result).

\subsubsection{LLM-based MG assistance capability evaluation method}
\begin{figure*}[t]
\begin{center}
\includegraphics[width=17cm]{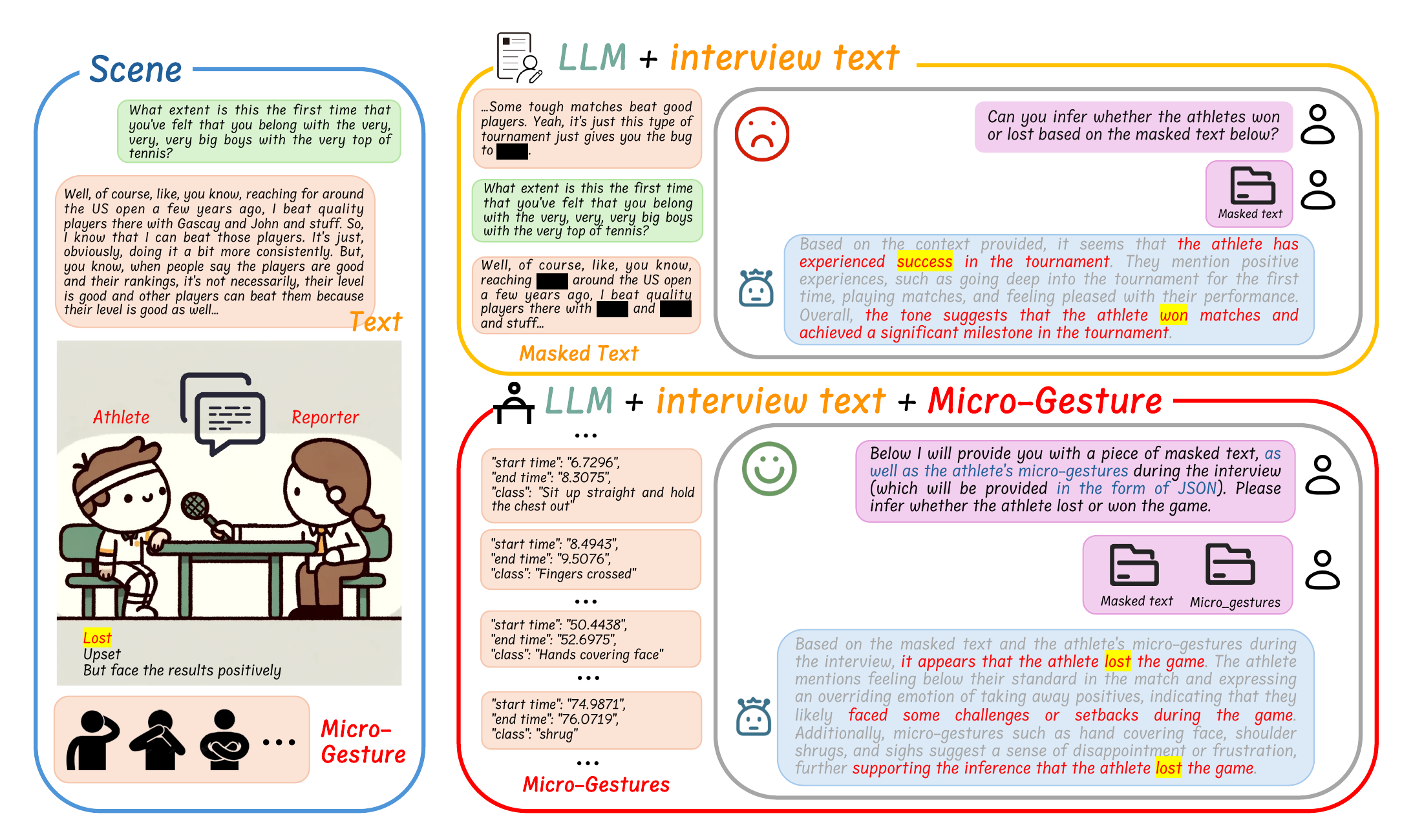}
\end{center}
\caption{An example of the validation of assistive role of micro-gestures in inferring emotion. The left blue box represents the constructed complex emotional reasoning scenario. When only the masked dialogue text is provided, the large language model (GPT-3.5) infers that the athlete won the match and achieved a milestone victory in the event. However, after inputting the micro-gesture information in JSON format, the model revises its previous inference. It then concludes that the athlete lost the match, with micro-gestures such as covering the face, shrugging, and sighing supporting the inference of the athlete feeling slightly dejected and disappointed.}
\label{fig:MGandEMOcase}
\end{figure*}

Based on the prior knowledge evaluation of LLMs, it is evident that most existing LLMs understand the objective outcomes of some match events. We have selected another modality of information, dialogue, to validate the supportive role of MGs in emotional inference. To minimize the leakage of information from the dialogue (such as the identity of the interviewee, venue conditions, interviewee's emotions, and match results), we have masked appropriate portions of the dialogue. Consider a complex scenario depicted in Fig.~\ref{fig:MGandEMOcase}, where the interviewee and the interviewer are engaged in a post-match interview discussing the just-concluded game and future plans. However, due to certain reasons, their conversation is intermittent, and you cannot access the entire dialogue, particularly noticing that some crucial information is lost in transmission. But you can ascertain which micro-gestures the interviewee made simultaneously. In such a scenario, can you infer the interviewee's current emotional state through the masked dialogue and micro-gestures indicating their actions? Existing large language models have demonstrated capabilities in text inference and detailed analysis that rival human performance. Therefore, we have tasked large language models with this challenge. The validation method comprises two key parts: dialogue processing and privacy protection. To minimize information breaches, and considering potential human biases in manual text processing, we provide a method utilizing GPT-3.5 to mask information. The prompt used is as follows.

``You are a professional text encryption expert. You will need to process the following text so that any information related to the outcome of the game and the player's identity is hidden within the text.
1) Please replace the text that needs to be hidden with [MASK].
2) Please process each sentence separately and output each sentence after hiding the key information according to the original format.
3) As an expert, you will not only hide information that directly reveals ``game results and player identity and emotions", but also hide clues that may reveal information, such as some emotion-related descriptors, or about some in-game situations).
The text you're dealing with is a post-match interview between a player and a reporter, a conversation between two people. Please organize the output into a Markdown table. The first column is the timestamp and the second column is the text behind the hidden information.
Every sentence must be strictly hidden, and words that may reveal the game situation in questions asked by reporters also need to be hidden. For example, words such as congratulations, winner, etc.''

Next, input the masked dialogues and micro-gesture information (in the form of JSON) into the LLMs. Prompt the LLMs to analyze and infer the athlete's emotion and the outcome of the match using only the masked dialogues and both of the dialogues and the micro-gesture information, at the same time LLMs are asked to provide a confidence level for each inference. The prompt format used is as follows.

``Below, I will send you the transcription of a post-match interview with an athlete, along with their micro-gestures during this process. Both types of information include timestamps, so you can match them up. First, please analyze only the text to determine the outcome of the athlete's performance in the match. Then, integrate the text with the micro-gesture information to analyze the outcome of the athlete's performance in the match. Please output the results in the following format: text-only: win: confidence score, lose: confidence score. text+micro-gestures: win: confidence score, lose: confidence score. Please indicate to what extent you think the athlete lost/won, and ensure that the sum of the confidence levels is 100 in both text-only and text+micro-gestures. When adding micro-gestures to the analysis, if clues indicate that the inference is contrary to the analysis using only text, please output it truthfully. You don't need to keep the results consistent between text-only analysis and micro-gesture analysis. Please do not output anything else.''
\section{Experiments}
\label{sec:exp}
\subsection{Micro-gesture recognition experiments}
\subsubsection{Datasets}
\label{sec:eva}
To better measure the effectiveness of our method, we not only tested it on the iMiGUE dataset. We also tested our model on two large action datasets: the NTU RGB+D 60 Action Dataset~\cite{shahroudy2016ntu} and the NTU RGB+D 120 Action Dataset~\cite{liu2019ntu}, and one small action dataset: Northwestern-UCLA Multi-view Action3D Dataset~\cite{wang2014cross}.

{\bf{iMiGUE.}}
To have standard evaluations for all the reported results on the iMiGUE dataset, two-level criteria have been defined. More specifically, on the MG recognition level, we utilized the cross-subject evaluation protocol which divides the 72 subjects into a training group of 37 subjects and a testing group of 35 subjects. The training and testing sets have 13\,936, and 4\,563 MG samples, respectively. The IDs of training and testing subjects can be found in the supplementary materials. On the emotion classification level, we selected 102 videos (51 wins and 51 loss matches) as the training set and 100 videos (50 win and 50 loss matches) as the test set. The player's emotional states (positive/negative) with the result of win or loss, are classified via analysis of MGs. The details of training and testing protocols (video IDs) can be found in the supplementary material. Especially, to benefit the community of skeleton or pose-based gesture recognition, we provide the pose data of every frame, achieved by using the OpenPose toolbox \cite{cao2017realtime}.

\begin{table*}
\centering
\caption{Linear evaluation results compared to SkeletonCLR and AimCLR on the NTU-60 and NTU-120 datasets. The ``$\Delta$'' in the ``SkeletonCLR'', ``AimCLR'' columns, and the first two rows of the ``Ours'' column represent the gain compared to the multistream fusion results. The ``$\Delta$'' in the last row of the ``Ours'' column represents the gain compared to the ``3S-AimCLR''. In the ``Ours+'' column, ``Joint+'' and ``3S'' use the encoder from AimCLR, and ``$\Delta$'' represents the gain compared to the Joint-stream results of AimCLR.}
 \vspace{-0.2cm}
\small
\tabcolsep2mm
\begin{tabular}{|l|l|cc|cc|cc|cc|}
\hline
\multirow{3}{*}{Method} & \multirow{3}{*}{Stream} & \multicolumn{4}{c|}{NTU-60(\%)} & \multicolumn{4}{c|}{NTU-120(\%)}                         \\ \cline{3-10}
   &     & \multicolumn{2}{c|}{xsub} & \multicolumn{2}{c|}{xview} & \multicolumn{2}{c|}{xsub} & \multicolumn{2}{c|}{xset} \\
   &     & acc.    & $\Delta$  & acc.    & $\Delta$     & acc.     & $\Delta$  & acc.     & $\Delta$       \\ \hline \hline
\multirow{4}{*}{SkeletonCLR \cite{li2021crossclr}}    & Joint              & 68.3   & $\uparrow$ 6.7  & 76.4    & $\uparrow$ 3.4      & 56.8   & $\uparrow$ 3.9  & 55.9   & $\uparrow$ 6.7\\
& Bone               & 69.4   & $\uparrow$ 5.6    & 67.4    & $\uparrow$  12.4    & 48.4   & $\uparrow$ 12.3 & 52.0   & $\uparrow$  10.6  \\
& Motion             & 53.3   & $\uparrow$ 21.7   & 50.8    & $\uparrow$ 29.0      &  39.6   & $\uparrow$ 21.1  & 40.2   & $\uparrow$ 22.4 \\
& 3S  & 75.0   & ——    & 79.8    & ——       & 60.7   &—— & 62.6   &——      \\ \hline
\multirow{4}{*}{AimCLR \cite{guo2022aimclr}} & Joint    & 74.3   & $\uparrow$ 4.6   & 79.7    & $\uparrow$ 4.1   & 63.4   & $\uparrow$ 4.8 &   63.4   & $\uparrow$ 5.4 \\ 
& Bone     & 73.2   & $\uparrow$ 5.7   & 77.0    & $\uparrow$ 4.1   & 62.9   & $\uparrow$ 5.3 & 63.4   & $\uparrow$ 5.4    \\
& Motion   & 66.8   & $\uparrow$ 12.1   & 70.6    & $\uparrow$ 13.2  & 57.3   & $\uparrow$ 10.9  & 54.4   & $\uparrow$ 14.4 \\
& 3S   & 78.9   &——   & 83.8    &——   & 68.2   &——   & 68.8   & ——  \\ \hline
\multirow{3}{*}{Ours} & Joint  & 72.16 & $\uparrow$ 7.72 & 71.27 & $\uparrow$ 15.17 & 58.86 & $\uparrow$ 9.94 & 61.97 & $\uparrow$ 7.47 \\
& Sequence  & 75.69 & $\uparrow$ 4.19 & 84.86 & $\uparrow$ 1.58 &58.44 & $\uparrow$ 10.36 & 58.09 & $\uparrow$ 11.35\\
& Joint+Sequence & \textbf{79.88} & $\uparrow$ 0.98 & \textbf{86.44} & $\uparrow$ 2.64 & \textbf{68.80} & $\uparrow$ 0.60 & \textbf{69.44} & $\uparrow$ 0.64 \\ \hline
\multirow{2}{*}{Ours\dag} & Joint\dag+Sequence & \textbf{80.93} & $\uparrow$ 6.63 & \textbf{87.60} & $\uparrow$ 7.9 &—— &  —— & —— &  ——\\
& 3S\dag+Sequence & \textbf{82.62} & $\uparrow$ 8.32 & \textbf{89.22} & $\uparrow$ 9.52 & —— &  —— & —— &  —— \\ \hline
\end{tabular}

\label{tb:gain}
\end{table*}

\begin{table}[t]
  \centering
  \small
  \caption{Ablation study of Temporal-Spatial-Balanced-Fusion on different datasets and protocols.}
  \vspace{0.2cm}
\begin{tabular}{|l|c|c|c|}
  \hline
  \multirow{2}{*}{Methods} & \multirow{2}{*}{iMiGUE} & \multicolumn{2}{c|}{NTU RGB+D} \\ \cline{3-4} 
                           &                          & C-View             & C-Sub             \\ 
  \hline \hline
  Joint                    & 41.05                    & 71.27              & 72.16             \\ 
  \hline
  Sequence                 & 35.22                    & 84.86              & 76.67             \\ 
  \hline
  Joint+Sequence           & 41.94                    & 86.44              & 79.88             \\ 
  \hline
\end{tabular}
\vspace{-0.2cm}
\label{tb:fusion}
\end{table}

{\bf{NTU RGB+D 60.}}
NTU RGB+D 60 is a mainstream public dataset in the field of action recognition. The dataset consists of 56\,880 samples, providing their RGB videos, depth maps, and skeleton data. A total of 40 subjects and 60 categories are included, and an actor is described by the 3D coordinates of 25 joints in the skeleton data. Two evaluation protocols are provided: 1) Cross-Subject (CS), where the actors in the training and test sets are different, with 40320 samples for training and 16560 samples for testing. 2) Cross-View (CV), where the camera angles in the training and test sets are different, with 37920 samples from two views (0° and 45°) for training and 18960 samples from one viewpoint (-45°) used for testing.

{\bf{NTU RGB+D 120.}}
The NTU RGB+D 120 dataset is an extension of NTU RGB+D 60 and is the largest benchmark for 3D action recognition. The dataset consists of 114,480 samples, providing their RGB videos, depth maps, and skeleton data. It contains 106 subjects, 32 settings, and 120 categories, giving an actor described by 3D coordinates of 25 joints in the skeleton data. Two evaluation protocols are provided: 1) Cross Setup (CSet), different camera distances and backgrounds make up different setups, where 54,471 samples with even setup numbers are used for training, and 59,477 samples with odd setup numbers are used for testing. 2) Cross Subjects (CS), consistent with NTU RGB+D 60, the actors in the training and testing sets are different, where 63,026 samples are used for training, and 50922 samples are used for testing.

\subsubsection{Experiment settings}
All experiments are conducted on the PyTorch deep learning framework on a PC with an Nvidia RTX 3090ti. All training configurations follow the original papers unless stated otherwise.

{\bf{Self-supervised Pretext training.}}
In the proposed plug-and-play temporal stream, we strive for model simplicity. To achieve this, we employ a contrastive learning network based on MOCO. The encoder is configured to comprise 3 layers of BI-GRU, each with 1024 units. In the spatial stream utilized during the experiments, we also employ a contrastive learning network based on MOCO. In this case, the encoder is chosen to be the joint stream in 2S-AGCN. During pre-training, the encoder is mapped to a 128-dimensional latent space embedding via the projection head but unattached during downstream tasks. Our training settings include a learning rate of 0.01, a decay factor of 0.0001, and a temperature value of $\tau=0.07$. We employ stochastic gradient descent (SGD) with Nesterov momentum (0.9) as the optimization strategy. In all our augmentation experiments, the network is trained until the loss converges and stabilizes. For the iMiGUE dataset under the cross-subject protocol, the size of the negative set N is set to 512. For NTURGB+D60, the size of the negative set N is set to 16,384.
 
{\bf{Linear Evaluation Protocol.}}
For a fair comparison, we follow the setup of previous work verifying the model through linear evaluation of action recognition tasks. This is how most contrastive learning models are evaluated. Specifically, we removed the projection head from the pre-trained query encoder since the projection head focuses on information about the pretext task, which is not the focus of the downstream task. We supervised trained a simple linear classifier (a fully connected layer followed by a softmax layer) with the encoder frozen. An SGD optimizer is used with a momentum of 0.9 and a learning rate of 0.1. A total of 100 epochs are trained by a linear classifier, and the learning rate decreased by a factor of 10 after the 50th and 80th epochs.
 
\subsubsection{Ablation Studies}
\label{sec:clip}
In this section, we perform ablation experiments on different datasets to verify the effectiveness of different components of our method. Specifically, we first analyze the effect of encoders with different bidirectional LSTM (BLSTM) layers on the results. We then delve into the effectiveness of the augmentation approach designed for micro-gesture in the paper and the comparison of the results for different combinations. Finally, we evaluate the results of using only single-dimensional and dual compensation.

{\bf{The effectiveness of the Temporal-Spatial-Balanced Dual-stream Fusion.}}
We conducted experiments on the iMiGUE, NTU RGB+D 60, and NTU RGB+D 120 datasets to test the effectiveness of the proposed ``Temporal-Spatial-Balanced Dual-stream Fusion'' method, and the results are shown in Table \ref{tb:fusion}. To further show that such a fusion method is more effective than the previous ``spatially heavy'' fusion method, the single-stream and fusion results were compared with SkeletonCLR and AimCLR on the NTU RGB+D 60 and NTU RGB+D 120 datasets in detailed comparison. The results are displayed in Table \ref{tb:gain}.

As can be seen in Table \ref{tb:fusion}, our method is effective for different datasets with different test protocols. The increase is significant when comparing fusion results with using only spatial streams (i.e., skeleton graph) and using only temporal streams (i.e., skeleton sequence). This verifies as we proposed in section \ref{sec:contrastive method} that the fusion of spatial and temporal streams can provide information gain for single-stream feature learning for MG recognition, jumping out from single-dimensional judgments and fusing multidimensional information to obtain more accurate results.

\begin{figure}[t]
\begin{center}
\includegraphics[width=8.3cm]{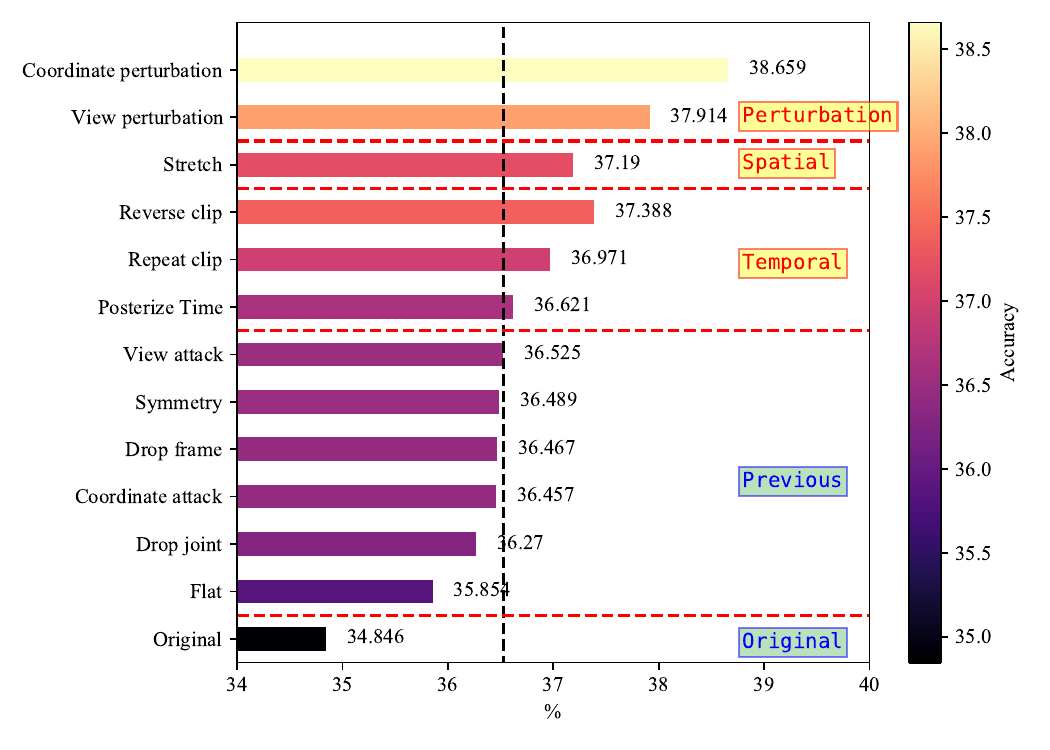}
\end{center}
\vspace{-0.42cm}
\caption{Linear evaluation results for single-strategy augmentations on iMiGUE (using joints only).}
\label{fig:augab1}
\end{figure}

Table \ref{tb:gain} lists the results of SkeletonCLR, AimCLR, and our method for single-stream and multi-stream fusion, respectively. SkeletonCLR and AimCLR adopt the fusion method of joint+bone+motion, which we call the ``spatially heavy'' fusion method in the previous section. As mentioned above, in our approach, we input the data to the GCN-based encoder in the form of Graphs (joints) and input data to the GRU-based encoder in the form of Sequences, respectively. We want to fuse the spatial structure features of the MGs learned by GCN and the temporal features of the motion learned by GRU. From this table, we can draw the following analytical conclusions:

i) Compared with the other two methods, the gain of our strategy for ``joint'' after fusion is the largest for each protocol in each dataset. This means that the ``temporal stream'' has more effect on the ``spatial stream'' than the ``spatial transformed stream'' in the fusion. It can provide additional information to help the model to get more comprehensive and accurate results.

ii) It is also worth noting that we directly use a GCN-based encoder in contrastive learning and use some simple augmentation methods. On the other hand, AimCLR has significantly better ``joint-only'' results than our method due to its ``extreme augmentation,'' ``nearest neighbor mining,'' and some other strategies. On NTU-60, it is 2.14\% higher under xsub and 8.43\% higher under xview. On NTU-120, it is 4.54\% higher under xsub and 1.43\% higher under xsetup. However, benefitting from our ``Temporal-Spatial-Balanced'' fusion method, as shown in the ``$\Delta$'' column in the last row of ``Ours'' column in Table \ref{tb:gain}, our final results are better than the ``joint-bone-motion'' streams fusion results.

iii) There is another interesting phenomenon that deserves our attention. Observing the penultimate row of the ``Ours'' column in Table \ref{tb:gain}, we can easily find that for the NTU-60 dataset, the temporal results are better than the spatial results, and we can assume that the temporal results dominate at this case. For the NTU-120 dataset, the temporal streams do not perform as well as the spatial streams, and the results for every single stream are lower than the ``joint'' streams of AimCLR. But after fusion, a very large gain is obtained for every single stream, and the fusion results are better than the three-stream fusion results of AimCLR. This is further evidence that the temporal and spatial streams have a strong complementary effect on each other. Since our fusion method focuses on time and space, such a balanced fusion approach outperforms the ``spatially heavy'' fusion of three streams when only two are used.

\begin{figure}[t]
\begin{center}
\includegraphics[width=8.3cm]{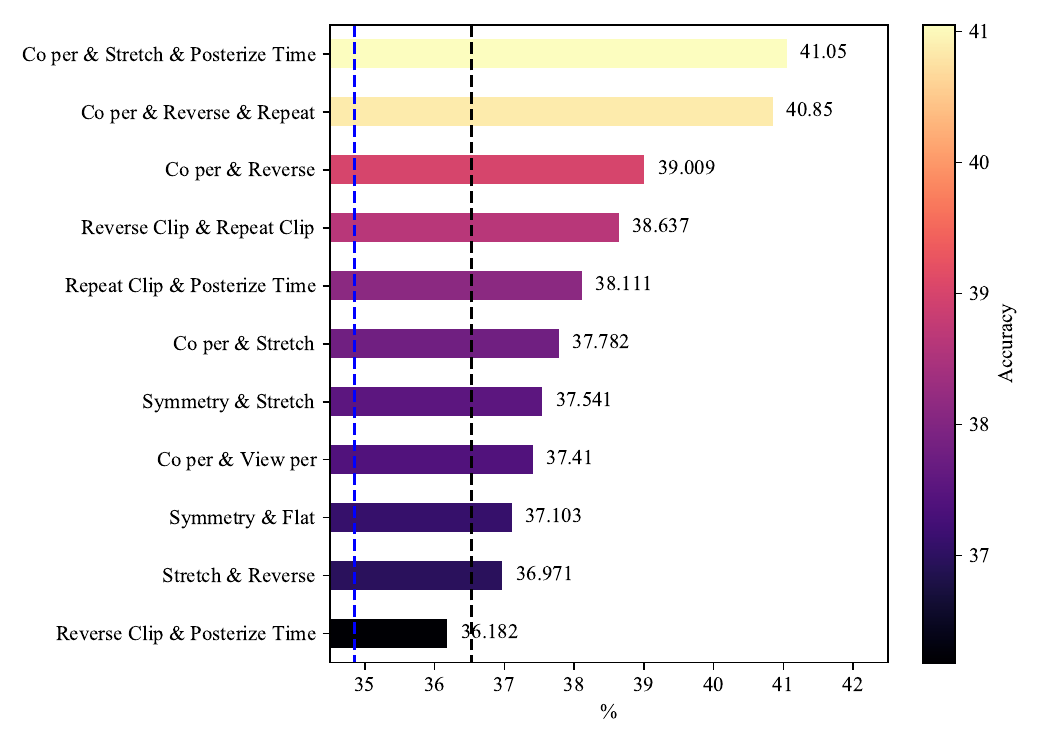}
\end{center}
\vspace{-0.42cm}
\caption{Linear evaluation results for multi-strategy augmentations on iMiGUE (using joints only).}
\label{fig:augab2}
\end{figure}

iv) In addition, in the ``Ours\dag'' column, we provide the results of the fusion of the better GCN encoder (from AimCLR) with our temporal stream (referred to as 2S in the following). This strongly demonstrates the effectiveness of our strategy: When comparing the gain of Joint streams in 3S- AimCLR, our 2S fusion has more gain with fewer streams fused (2.03\% higher for xsub and 3.8\% higher for xview in NTU-60). We also provide 3S and temporal stream fusion results, which are much higher than those of 3S fusion. This proves that our temporal streams provide sufficient information that was not exploited before.

{\bf{The effectiveness of the MG augmentations on MG recognition.}} To systematically evaluate the effectiveness of our proposed data augmentation strategy for micro-gesture recognition, we tested the performance of different augmentation strategies and their various combinations on the iMiGUE dataset.

We first show in Fig.~\ref{fig:augab1} the gain of a single MG augmentation approach helping the model tap into the potential consistency of different actions. We give results not only for approaches proposed in previous action recognition work (including {\emph{coordinate attacks}}, {\emph{viewpoint attacks}}, {\emph{drop node}}, {\emph{symmetry}}, and {\emph{drop frames}}), but also for our augmentation approaches for MG (temporal: {\emph{reverse clip}}, {\emph{repeat clip}}, and {\emph{posterize time}}; spatial: {\emph{stretch}}; perturbations: {\emph{coordinate perturbation}}, {\emph{viewpoint perturbation}}).

Fig.\ref{fig:augab1} is divided into five panels by four horizontal red dashed lines, which correspond to their augmentation. ``Original'' represents the results obtained without any augmentation, and ``Previous'' shows the augmentations proposed in previous work for general action recognition. The remaining three categories correspond to the methods proposed in Sec. \ref{sec:augmentation method}. The results are arranged from highest to lowest in each of the five panels. The vertical black dashed line represents the maximum value that previous augmentation can obtain. From this diagram, we can draw the following analysis and conclusions:

i) Without using any augmentation method, i.e., directly using the original skeleton graph, we obtain a result of 34.85\%, which is better than the results shown in Table \ref{tb:mgreg} for both P\&C (31.67\%) and U-S-VAE Z (32.43\%). This indicates that our proposed DoCLR is more effective for learning the original skeleton graph representations. The model can focus more on the uniqueness of each instance and its difference from the others, and obtain better edge features of the instances in the process of ``pushing away'' the negative samples. This is a significant advance over methods that focus only on single instance feature extraction.

ii) Applying different augmentation strategies in DoCLR has a positive gain in the learning of the representations compared to using the original graph directly. The lowest score with {\emph{flat}} in the previous methods (35.85\%) outperforms the results obtained by AS-CAL using the optimal augmentation combination (35.48\%) shown in Table \ref{tb:mgreg}. This indicates that our method has enhanced feature extraction capability compared to AS-CAL. The result is improved by almost 4 percent when using ``coordinate perturbation''. This validates that our augmentation methods can increase the difficulty level of the contrastive learning task thus encouraging the model to learn better representations.

iii) Compared to the previously proposed methods, the new approach achieves superior performance on the micro-gesture dataset, with all the results of the method exceeding the previous optimal value (36.53\%). This shows that the augmentation approach designed for micro-gesture recognition is more efficient in generating different representations of the reality of a micro-gesture without destroying the original category, thus learning more ``different'' representations of the same action.

\begin{table}
\centering
\caption{Comparison of MG recognition accuracy (\%) with state-of-the-art algorithms on the iMiGUE dataset (best: bold, second best: underlined).}
\footnotesize
\resizebox{0.48\textwidth}{!}{\begin{tabular}{|c|l|c|c|c|}
\hline
  \multicolumn{2}{|c|}{\multirow{2}{*}{\tabincell{l}{Methods}}} & \multirow{2}{*}{\tabincell{l}{Model+Modality}} & \multicolumn{2}{c|}{iMiGUE}\\\cline{4-5}
  \multicolumn{2}{|c|}{}  &  & Top-1 & Top-5\\
    \hline \hline
 \multirow{15}{*}{\tabincell{l}{Super-\\vised}}     &      S-VAE \cite{shi2018bidirectional}   &   \multirow{3}{*}{\tabincell{l}{RNN + Pose}}        & 27.38 & 60.44\\
      &      LSTM                                &                               & 32.36 & 72.93\\
      &      BLSTM                               &                               & 32.39 & 71.34 \\
    \cline{2-5}
      &      ST-GCN \cite{yan2018spatial}        &  \multirow{5}{*}{\tabincell{l}{GCN + Pose}}         & 46.97 & 84.09\\
      &      2S-GCN \cite{shi2019two}            &                               & 47.78 & 88.43\\
      &      Shift-GCN \cite{cheng2020skeleton}  &                               & 51.51 & 88.18\\
      &      GCN-NAS \cite{peng2020learning}     &                               & 53.90 & 89.21\\
      &      MS-G3D \cite{liu2020disentangling}  &                               & 54.91 & \underline{89.98}\\
    \cline{2-5}
      &      C3D \cite{tran2015learning}	        &	\multirow{3}{*}{\tabincell{l}{3DCNN + RGB}}	 & 20.32  	& 55.31	\\
      &      R3D-101  \cite{hara2018can}		    &		                     & 25.27     & 59.39 \\
      &      I3D  \cite{carreira2017quo}         &                               & 34.96     & 63.69 \\
    \cline{2-5}
      &      TSN \cite{wang2018temporal}         &   \multirow{3}{*}{\tabincell{l}{2DCNN + RGB}}     & 51.54  	& 85.42 \\
      &      TRN \cite{zhou2018temporal}         &                              & \underline{55.24}  	& 89.17	\\
      &      TSM \cite{lin2019tsm}               &                               & \textbf{61.10}  	& \textbf{91.24} \\
    \hline
\multirow{4}{*}{\tabincell{l}{Unsup-\\ervised}}      &      P\&C \cite{su2020predict}                   &   \multirow{2}{*}{\tabincell{l}{Encoder-Decoder\\+ Pose}}             & 31.67 & 64.93\\
      &      U-S-VAE Z \cite{Liu_2021_CVPR}                            &                                     & 32.43 & 64.30\\
      \cline{2-5}
      &      AS-CAL \cite{Rao2021AugmentedSB}             &   \multirow{2}{*}{\tabincell{l}{Contrastive Learning\\+Pose}}             & \underline{35.48} & \underline{75.32}\\
      &      \textbf{Ours}                   &  & \textbf{41.94}  &\textbf{81.6}\\
   \hline
\end{tabular}}
\setlength{\abovecaptionskip}{0pt}

\vspace{-0.4cm}
\label{tb:mgreg}
\end{table}

iv) Among the proposed augmentations, the best class is the ``perturbation'' augmentation. The temporal augmentations are designed according to the ``repetitiveness'' and ``randomness'' of the micro-gesture in the temporal order, which can increase the semantic information of the action to some extent. The ``perturbation'' type of augmentation, on the other hand, not only adds appropriate noise to the MG but also increases the probability of a more novel representation of action due to the perturbation of coordinate and viewpoint.

We then show in Fig.~\ref{fig:augab2} the benefits of augmentation of two or three (by our proposed method) combinations of MG in graph-based representation learning for MG recognition. We tried combinations within the same category (e.g., Reverse Clip combined with Repeat Clip), combinations within different categories (e.g., Co-per combined with Reverse), and combinations across multiple categories (e.g., Co-per combined with Stretch and Posterize Time).

The vertical black dashed line in Fig.~\ref{fig:augab2} has the same meaning as in Fig.~\ref{fig:augab1} (i.e., 36.63\%), and the vertical blue dashed line represents the result obtained without any augmentation (i.e., 34.85\%). For this figure, we can obtain the following analysis and conclusions.

i) The combination strategy applies several different augmentations to the skeletal graph to further increase the complexity of the skeletal semantic information. It can be seen that the combination strategy can improve the performance of our method, and some combinations are significant for the performance improvement, such as Coordinate Perturbation combined with Stretch and Posterize Time can improve 6.20\% compared with the original. This shows that a reasonable combination of strategies can further improve the recognition results.

ii) When Reverse Clip was combined with Posterize Time, the results were not even as good as the previous method, although there was also an improvement compared to the original. It can be inferred that such a combination of sampling at different speeds during action repetition is likely to lose information about key action frames, thus affecting the final results. Similarly, Coordinate perturbation was able to achieve a result of 38.60\% when used alone but only obtained a result of 37.41\% when combined with view perturbation. It can be inferred that simultaneous changes in joints and viewpoints of the actions can cause drastic changes in itself, thus substantially increasing the difficulty of recognition.

iii) The combination strategy of Coordinate perturbation with Stretch and Posterize Time works better than any other method. This combination takes into account the three approaches in perturbation, temporal and spatial. The model is encouraged to resist the noise of joint point movement better while being able to learn consistent semantics implied by the same action under different body features and movement habits. This allows the model to learn richer features and obtain superior results.

\begin{figure}[t]
\begin{center}
\includegraphics[width=8.3cm]{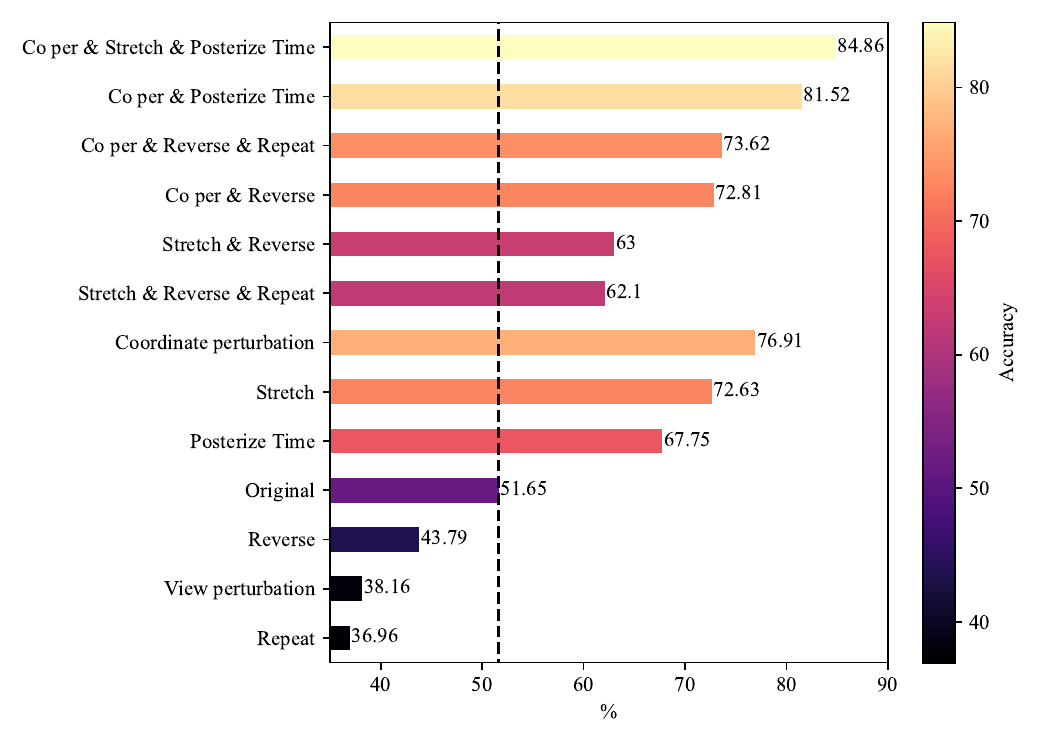}
\end{center}
\vspace{-0.42cm}
\caption{Linear evaluation results for single- and multi-strategy augmentations on NTU-60 (using sequence only).}
\label{fig:augab3}
\end{figure}

{\bf{The effectiveness of the MG augmentations on general action recognition.}}
We linear evaluated the augmentation approaches designed for MGs on NTU-60 which is a general action dataset, and the results are shown in Fig.~\ref{fig:augab3}. We present both the results for single-strategy augmentations and multi-strategy combination augmentations. As in Fig. b, the vertical blue dashed line in Fig.~\ref{fig:augab2} represents the result without any augmentation, i.e., 51.65\%. For this figure, we can obtain the following analysis and conclusions:

\begin{figure}[t]
\begin{center}
\includegraphics[width=8.3cm]{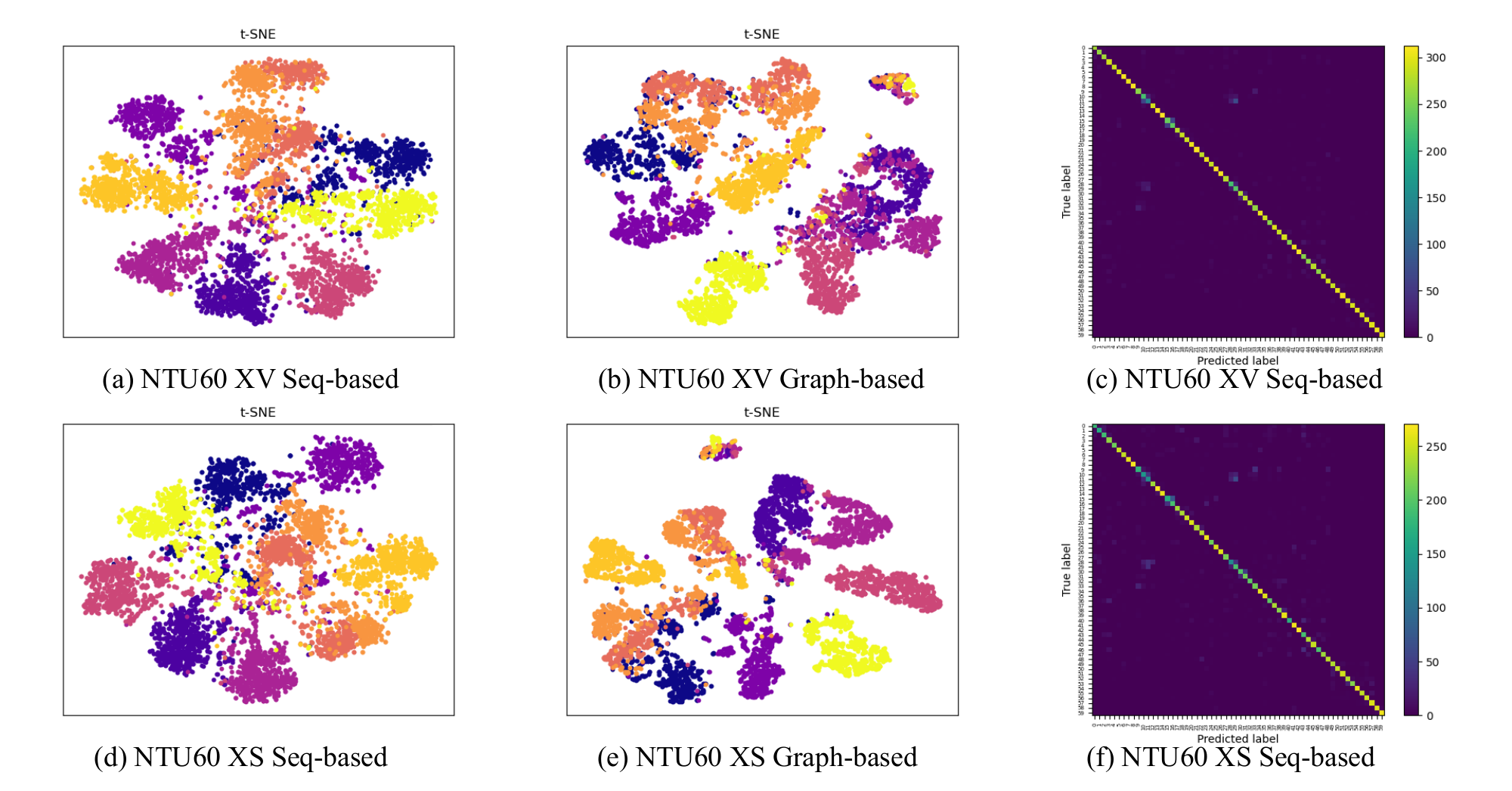}
\end{center}
\vspace{-0.42cm}
\caption{T-SNE visualization and confusion matrix on NTU-60. XS: cross-sub; XV: cross-view.}
\label{fig:TSNE}
\end{figure}

i) Not all MG augmentations are applicable to general actions. It can be seen that ``repeat'', ``reverse'', and ``view perturbation'' do not have a positive effect. Among them, ``repeat'' and ``reverse'' are specifically proposed for the possible temporal repetition of MGs, which is rare but not absent (e.g., brushing teeth) in general actions. Therefore, when such augmentation is applied to a general action, it destroys the original continuous temporal semantic information, thus having a negative effect.

ii) Some strategies still work, such as ``coordinator perturbation'', ``stretch'', and some combinations of strategies. Most of these strategies still make sense for the semantics of general actions. For example, for ``stretch'', the subjects of the general actions also have different movement habits and body types, which are the same as the characteristics of MGs. Similarly, for ``posterize time'', most people have their own appropriate speed whether they are doing general actions or MGs. It is because MGs and general actions have some common characteristics, so some strategies can show good performance.

\subsubsection{Comparison with state-of-the-art methods}
\label{sec:video}

{\bf{Linear Evaluation Results on iMiGUE.}}
In order to evaluate supervised learning-based methods' performance on iMiGUE, 14 state-of-the-art algorithms are selected which can be simply categorized into four groups, namely, body key-points-based RNN (\textit{i}.\textit{e}., BLSTM, LSTM, and S-VAE \cite{shi2018bidirectional}), and GCN (\textit{i}.\textit{e}., ST-GCN \cite{yan2018spatial}, 2S-GCN \cite{shi2019two}, Shift-GCN \cite{cheng2020skeleton}, GCN-NAS \cite{peng2020learning}, and MS-G3D \cite{liu2020disentangling}), RGB-based 3DCNN (\textit{i}.\textit{e}., C3D \cite{tran2015learning}, R3D-101 \cite{hara2018can}, and I3D \cite{carreira2017quo}), and 2DCNN with temporal reasoning (\textit{i}.\textit{e}., TSN \cite{wang2018temporal}, TRN \cite{zhou2018temporal}, and TSM \cite{lin2019tsm}). We further evaluate the effectiveness of the proposed method by comparing it with existing unsupervised methods on iMiGUE. Here, we report the results of P\&C \cite{su2020predict} and AS-CAL \cite{Rao2021AugmentedSB} on iMiGUE since their implementation code is public and they are friendly to equipment and training time. It is noted that all models follow the same evaluation protocol mentioned above for a fair comparison. In Table \ref{tb:mgreg}, we present the performances of these baseline networks.

\begin{figure}[t]
\begin{center}
\includegraphics[width=8.3cm]{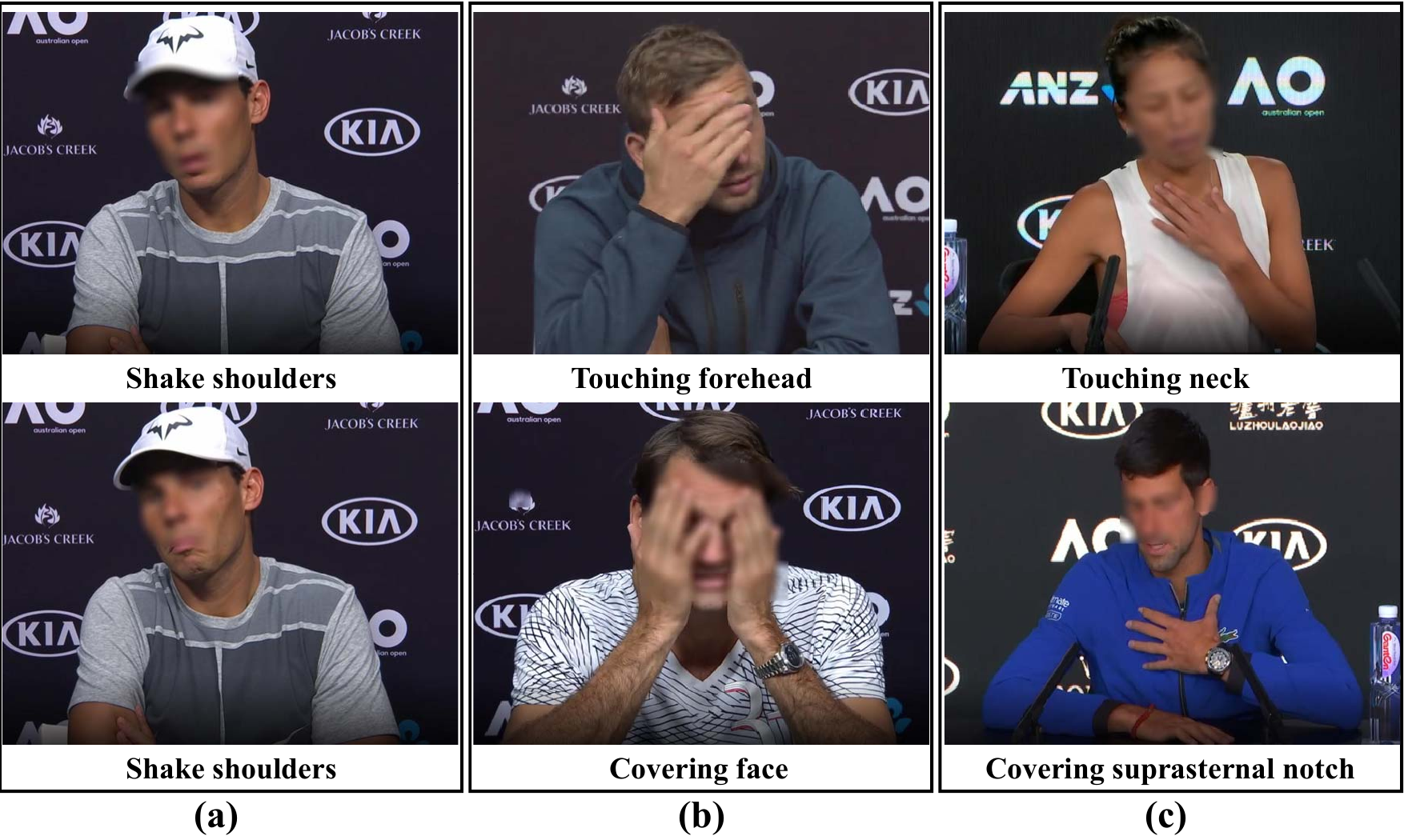}
\end{center}
\vspace{-0.42cm}
\caption{Examples of challenging recognition of the micro-gestures.}
\label{fig:similar}
\end{figure}

From Table \ref{tb:mgreg}, we can summarize several observations: 1) Almost all of methods' accuracy (Top-1) stuck under 60 percentage, which could verify that recognizing such hardly noticeable MGs (\textit{e}.\textit{g}., a short-timing ``Shake shoulders'' as shown in Fig. \ref{fig:similar} (a)) is a very challenging task. Due to the subtle differences between MGs (\textit{e}.\textit{g}., ``Covering face'' vs. ``Touching forehead'', ``Touching neck'' vs. ``Covering suprasternal notch'' as shown in Fig. \ref{fig:similar} (b) and (c)), visual or structural appearances in the form of RGB or pose contribute significantly less than that in a regular gesture (action) recognition task. 2) 3DCNN and RNN-based models' Top-1 performance is lower than 35$\%$, which is not surprising as fully-supervised learning models may have a significant performance drop with class imbalance issue. 3) Capturing temporal dynamics (temporal reasoning) is important as 2DCNN-based TSM and TRN outperform others by large margins. 4) Our method outperforms all prior unsupervised learning models. Although not using any labels, our performance is very competitive with the supervised 3DCNN and RNN-based methods.

\begin{table}
\centering
\caption{Comparison of action recognition accuracy (\%) with state-of-the-art algorithms on the NTU RGB+D 60 dataset (best: bold, second best: underlined).}
\footnotesize
\begin{tabular}{|l|c|c|c|}
\hline
    \multirow{2}{*}{\tabincell{l}{Unsupervised Methods}}  & \multirow{2}{*}{\tabincell{l}{Modality}} & \multicolumn{2}{c|}{NTU RGB+D}\\
    \cline{3-4}
    & & XView & XSub\\
    \hline \hline
      Shuffle \& Learn \cite{misra2016shuffle}   & \multirow{4}{*}{\tabincell{l}{RGB-D}}   & 40.9 & 46.2\\
      Luo \textit{et al.} \cite{luo2017unsupervised} &   & 53.2 & 61.4\\
      Li \textit{et al.} \cite{li2018unsupervised} &  & 63.9 & 68.1 \\
    \hline

      LongT GAN \cite{zheng2018unsupervised}& \multirow{13}{*}{\tabincell{l}{pose}} & 48.1 & 39.1 \\
      P\&C (CVPR $20$) \cite{su2020predict} &   & 76.1 & 50.7\\
      U-S-VAE (CVPR $21$) \cite{Liu_2021_CVPR} &    & 64.9 & 51.0\\
      MS$^{2}$L (ACM MM $20$) \cite{10.1145/3394171.3413548} &   & - & 52.6\\
      AS-CAL (Info. Sciences $21$) \cite{Rao2021AugmentedSB} &  & 64.8 & 58.5\\
      SeBiReNet (ECCV $20$) \cite{nie2020unsupervised} &    & 79.7 & -\\
      AimCLR (AAAI $22$) \cite{guo2022aimclr} & &79.7 &74.3\\
      TAHAR (WACV $24$) \cite{Lerch_2024_WACV} & & 82.1 & 72.4\\
      ISC (ACM MM $21$) \cite{10.1145/3474085.3475307} &  & \underline{85.2} & 76.3 \\
      3s-SkeletonCLR (CVPR $21$) \cite{li2021crossclr} &    & 79.8  & 75 \\
      3s-Colorization (ICCV $21$) \cite{yang2021skeleton} &   & 83.1 & 75.2 \\
      3s-CrosSCLR (CVPR $21$) \cite{li2021crossclr} &  & 83.4 & 77.8 \\
      3s-AimCLR (AAAI $22$) \cite{guo2022aimclr} &   & 83.8 & 78.9 \\
      3s-RVTCLR+ (ICCV 23) \cite{zhu23Modeling} & & 84.6 & \underline{79.7}\\
      3s-HYSP (ICLR 23) \cite{franco2023hyperbolic} & & \underline{85.2} & 79.1\\
      \textbf{Ours (Two Stream)} &   & \textbf{86.4} & \textbf{79.9}\\
      \hline
      3s-HCCL (AAAI $23$) \cite{zhang2023hierarchical} & \multirow{5}{*}{\tabincell{l}{pose}} & 85.5 & 80.4\\
      3s-SkeleMixCLR \cite{chen2022contrastive} & & 87.1 & 82.7 \\  
      3s-CPM (ECCV $22$)\cite{zhang2022contrastive} & &87.0 & \underline{83.2}\\
      3s-ActCLR (CVPR $23$) \cite{lin2023actionlet} & & \underline{88.8} & \textbf{84.3}\\
      \textbf{Ours\dag} & & \textbf{89.2} & 82.6\\
      \hline
\end{tabular}
\setlength{\abovecaptionskip}{0pt}
\vspace{-0.4cm}
\label{tb:ntureg}
\end{table}

{\bf{Linear Evaluation Results on NTU-60.}} 
To test the generalization capability of our method, we provide its performance on different datasets, such as the NTU RGB+D 60 \cite{shahroudy2016ntu}. Not merely provided the commonly cross-subject (XSub) protocol, the authors of NTU RGB+D also recommended the cross-view (XView) evaluation. We follow this convention and report the recognition accuracy (Top-1) of the two protocols. Here, the comparison results of RGB-D-based unsupervised methods 
and pose-based unsupervised methods 
are presented.

As shown in Table \ref{tb:ntureg}, our method outperforms other comparative single-stream (i.e., joint stream) and most three-stream (i.e., joint, bone, and motion stream) fusion methods when using only spatiotemporal stream fusion. Compared to the 3s-HYSP method, our method leads by 1.2\% and 0.8\% under the XView and XSub protocols, respectively, and outperforms 3S-AimCLR by 2.64\% and 0.98\%, and ISC by 1.24\% and 3.58\%. It is worth mentioning that the ISC and three-stream methods utilize cross-stream knowledge mining strategies, providing additional positive and negative sample information through other streams. Moreover, when 3S-AimCLR and 3s-HYSP use three streams for fusion, our method achieves superior results using only two streams. This indicates that our ``spatio-temporal balanced fusion'' is effective, as incorporating temporal stream information can provide more information gain to the joint stream. Our fusion method enables the model to access temporal features that cannot be obtained solely through exploring spatial structures. Additionally, by further fusing more complex spatial streams (from \cite{guo2022aimclr}) and temporal streams (consistent with those in ``Ours''), our method can surpass almost all other comparative methods under the XView and XSub protocols.

\begin{table}[htp]
\centering
\caption{Comparison of action recognition accuracy (\%) with state-of-the-art algorithms on the NTU RGB+D 120 dataset (best: bold, second best: underlined).}
\small
\begin{tabular}{|l|c|c|}
\hline
Method                            & XSub(\%)       & XSet(\%)        \\      \hline
    \hline
P\&C (CVPR $20$) \cite{su2020predict}    & 42.7           & 41.7            \\
AS-CAL (Info. Sciences $21$) \cite{Rao2021AugmentedSB} & 48.6           & 49.2            \\
TAHAR (WACV $24$) \cite{Lerch_2024_WACV} & 65.1 & 64.8\\
3s-CrosSCLR $^\S$ (CVPR $21$) \cite{li2021crossclr}      & 67.9           & 66.7            \\
ISC (ACM MM $21$) \cite{10.1145/3474085.3475307}  & 67.9      & 67.1            \\
3s-AimCLR (AAAI $22$) \cite{guo2022aimclr}    & \underline{68.2}  & 68.8   \\ 
3s-RVTCLR+ (ICCV 23) \cite{zhu23Modeling} & 68.0 & \underline{68.9}\\
3s-HYSP (ICLR 23) \cite{franco2023hyperbolic}& 64.5 & 67.3\\
\textbf{Ours (Two Stream)} & \textbf{68.8} & \textbf{69.4} \\ 
    \hline
\end{tabular}
\vspace{-0.4cm}
\label{tb:ntu120-com}
\end{table}

\begin{table}[htp]
  \centering
  \caption{Comparison of emotion understanding accuracy (\%) with methods on the iMiGUE dataset (best: bold, second best: underlined).}
   \footnotesize
    \begin{tabular}{|l|c|c|}
    \hline Methods & Model + Modality & Accuracy\\
    \hline \hline
            TRN \cite{zhou2018temporal}        & \multirow{3}{*}{\tabincell{l}{CNN + RGB}}                & 0.53 		\\
            TSM \cite{lin2019tsm}              &        & 0.53  	\\
            I3D \cite{carreira2017quo}         &                 & 0.57     \\
    \hline
            ST-GCN \cite{yan2018spatial}       &  \multirow{2}{*}{\tabincell{l}{GCN + Pose}}           & 0.50 \\
            MS-G3D \cite{liu2020disentangling} &          & 0.55 \\
    \hline
            U-S-VAE \cite{Liu_2021_CVPR} + LSTM                     & \multirow{2}{*}{\tabincell{l}{RNN + Pose}}         & 0.55 \\
            TSM \cite{lin2019tsm} + LSTM       &          & \underline{0.60} \\
     \hline
                 GPT3.5 \cite{achiam2023gpt} & LLM + Text + MG         & \textbf{0.66} \\
     \hline
    \end{tabular}

\label{tb:emoreg}
\end{table}%

{\bf{Linear Evaluation Results on NTU-120.}} 
We report the recognition accuracy (top-1) of our method under cross-subject (XSub) and cross-setup (XSet) protocols on NTU-120. Table \ref{tb:ntu120-com} shows that our method beats other comparative self-supervised methods on NTU-120. Our fusion results outperformed the advanced 3S-AimCLR (68.8\% vs. 68.2\% on XSub and 69.4\% vs. 68.8\% on XSet); 3s-RVTCLR+ (0.8\% on XSub and 0.5\% on XSet); 3s-HYSP (4.3\% on XSub and 2.1\% on XSet). Such comparative results are produced under the condition that we use only two stream fusions while 3S-AimCLR uses three stream fusions. This shows that our method is equally competitive on large-scale datasets.

\subsection{MG-based Emotion Understanding Experiments}
\subsubsection{MG-based-only Emotion Understanding}

\begin{figure*}[t]
\begin{center}
\includegraphics[width=17cm]{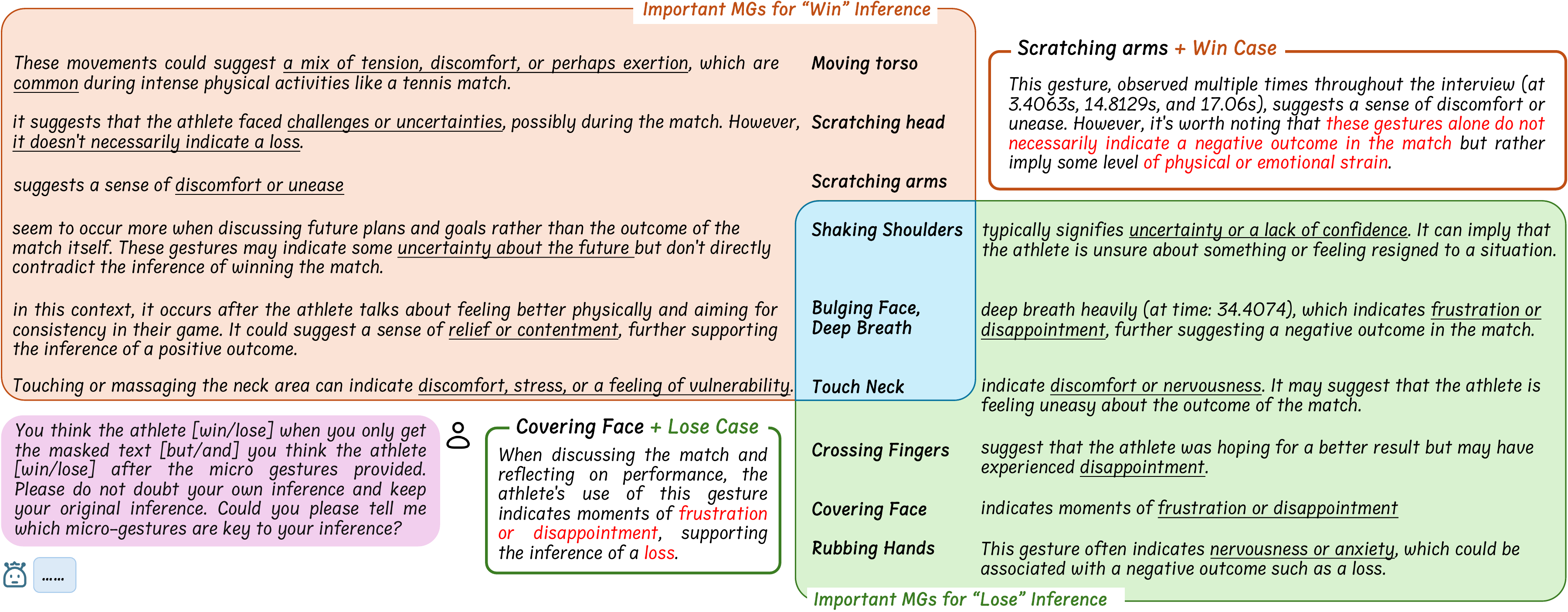}
\end{center}
\caption{The most frequently occurring micro-gestures deemed pivotal in LLM judgments.}
\label{fig:top-MG}
\end{figure*}

\begin{table*}[htp]
    \centering
    \caption{Comparison of method performance}
    \begin{tabular}{|c|c|c|c||c|c|c|}
        \hline
        \multirow{2}{*}{Method} & \multicolumn{3}{c||}{GPT3.5\cite{achiam2023gpt}} & \multicolumn{3}{c|}{Gemini Pro \cite{team2023gemini}} \\
        \cline{2-7}
         & Acc@1 & Acc@3 & Acc@5 & Acc@1 & Acc@3 & Acc@5 \\
        \hline
        Masked Text-Only & 60.44 & 58.24 & 57.7 & 61 & 59.67 & 58.8 \\
        Masked Text + MG & 67.03 & 67.4 & 65.81 & 64 & 62.67 & 61 \\
        \hline
    \end{tabular}
    \label{tab:LLM_EMO}
\end{table*}

In our previous research\cite{Liu_2021_CVPR}, we conducted a quantitative analysis of emotion understanding based solely on micro-gestures, to evaluate the performance of state-of-the-art methods in video-level emotion understanding tasks, all based on RGB or skeleton data. Among these models, I3D achieved the highest score of 60\% (see Table \ref{tb:emoreg}). It combines outputs of multiple parallel branches at the same level but different resolutions, offering a richer representation of emotion classification. Our results indicate that micro-gestures can, to some extent, be used independently for emotion inference. 

\subsubsection{MG-as-assistant Emotion Understanding}
However, in field environments, micro-gestures rarely occur alone, and psychologists seldom rely on them as isolated cues. Compared to acting as standalone cues, micro-gestures might be more effective in a supporting role. Therefore, we employed a large language model—a ``psychologist'' with extensive and deep knowledge—to conduct this experiment. In Section \ref{sec:LLM_Prior}, we tested the large language model's prior knowledge of micro-gestures and set up a complex evaluation scenario. In this scenario, emotions and match information in the conversation between athletes and reporters were masked to ensure no leakage of actual outcomes. The large language model was encouraged to integrate micro-gestures with the dialogue for comprehensive inference, analyzing the emotional tone of the athletes during the game. We report the inference results based on masked dialogue alone and in combination with micro-gestures in Table \ref{tab:LLM_EMO}, repeating the experiment five times to ensure the stability of the results. Notably, both the dialogue and micro-gestures were input in text form, with no use of any RGB modalities. The results show that our masking strategy effectively prevented information leaks, hindering the model from deducing outcomes directly from clues in the dialogue. With the addition of micro-gesture information, for GPT3.5, the inference accuracy improved by at least 6.59\%; for Gemini Pro, it increased by at least 2.2\%. Gemini Pro and GPT3.5 are not from the same family of large language models, which somewhat ensures that the improvement is not merely due to model training biases. Moreover, this evaluation experiment achieved better results than previous trained models based on microgestures in a zero-shot condition, reaching an average accuracy of 65.81\% in GPT3.5 tests. This demonstrates that micro-gestures, as auxiliary cues, can positively guide emotion inference, which is significant for our further exploration of micro-gestures. Imagine scenarios such as deceptive conversations and interviews hiding true situations; even if information in other modalities is deliberately masked or misleading, micro-gestures might still reveal true thoughts to some extent.

When using large language models to infer emotions from redacted text combined with micro-gestures, some ``situation reversal'' cases occurred. In these cases, the inference based solely on the masked text was inconsistent with the factual baseline, but when micro-gesture information was provided, the correct outcome was achieved. In these instances, the large language model explained which specific micro-gestures influenced it to revise its previous inference. As shown in Fig.~\ref{fig:top-MG}, we collected these micro-gestures and displayed the nine most frequently mentioned by the large language model. For example, moving torso, scratching head, and scratching arms played an important role in inferring positive emotions, while Crossing fingers, Covering face, and Rubbing hands were crucial for inferring negative emotions. Additionally, shaking shoulder, bulging face/deep creath, and touching neck provided extra judgment information to the large language model, but their impact varied depending on the combination of micro-gestures and the textual context, serving as important clues for both positive and negative emotions. To further understand and ascertain the role of these micro-gestures, the large language model also provided detailed explanations of the emotions represented by these gestures, revealing that some gestures commonly associated with both positive and negative emotions often signify nervousness, anxiety, or stress relief.
\section{Conclusion}
This paper builds on iMiGUE to further advance the study of micro-gestures. The work goes beyond merely examining representative methods for MG recognition, and instead explores the connection between MGs and the understanding of emotional states. Considering that current unsupervised action recognition methods based on skeletons still use a ``space-heavy'' fusion form, we propose a spatio-temporal balanced dual-stream contrastive learning network. We have also designed new augmentation methods for MG attributes to drive the model to learn more intrinsic consistencies. We evaluated the capability of mainstream large language models in understanding micro-gestures and validated that micro-gestures can play a positive role in aiding emotional understanding through large language models. We hope these efforts will contribute to new advancements in the field of emotional artificial intelligence. In the future, we will explore more multimodal cues related to micro-gestures and investigate complex models to more comprehensively find the potential mappings between emotions and MGs.

\section*{Data Availability}
The datasets generated during and/or analyzed during the current study are available at: \href{https://github.com/Erich-G/MG_based_Emotion_Understanding}{https://github.com/Erich-G/MG\_based\_Emotion\_Understanding}. 



\bibliographystyle{IEEEtran}
\bibliography{main}

\end{document}